\documentclass[conference]{IEEEtran}
\usepackage{times}

\usepackage[numbers]{natbib}
\usepackage{multicol}
\usepackage[bookmarks=true]{hyperref}
\usepackage{amsmath}
\usepackage{amsfonts}
\usepackage{amssymb}
\usepackage{bm}
\usepackage{calligra}
\usepackage{subcaption}
\usepackage{breqn}
\usepackage{sidecap}
\usepackage{balance}
\usepackage{textcomp}
\usepackage{float}
\usepackage{graphicx}

\newsavebox{\twosubbox}
\usepackage{url}
\usepackage[font=footnotesize, skip=5pt]{caption}

\newcommand{\x}{\bm{x}} 
\newcommand{\z}{\bm{z}} 

\newcommand{\Metric}{\bm{M}} 
\newcommand{\curve}{c} 
\newcommand{\Jac}{\bm{J}} 
\newcommand{\trsp}{\mathsf{T}} 

\newcommand{\ambient}{\mathcal{X}} 
\newcommand{\latent}{\mathcal{Z}} 
\newcommand{\Manifold}{\mathcal{M}} 
\newcommand{\R}{\mathbb{R}} 
\newcommand{\Sph}{\mathcal{S}} 
\newcommand{\Q}{\mathcal{Q}} 

\newcommand{\I}{\mathbb{I}} 
\newcommand{\Prob}{p} 
\newcommand{\vMF}{\mathrm{vMF}}
\newcommand{\Length}{\mathcal{L}} 
\newcommand{\q}{\bm{q}} 
\newcommand{\Loss}{\mathcal{L}} 

\newcommand{\g}{\bm{g}} 

\begin{document}

\title{Learning Riemannian Manifolds for \\Geodesic Motion Skills}
\author{\authorblockN{Hadi Beik-Mohammadi$^{1,2}$, S\o{}ren Hauberg$^{3}$, Georgios Arvanitidis$^{4}$, Gerhard Neumann$^{2}$, and Leonel Rozo$^{1}$}
\authorblockA{$^{1}$\mbox{Bosch Center for Artificial Intelligence (BCAI)}, Renningen, Germany.\\
$^{2}$ Autonomous Learning Robots Lab, Karlsruhe Institute of Technology (KIT), Karlsruhe, Germany.\\
$^{3}$ Section for Cognitive Systems, Technical University of Denmark (DTU), Lyngby, Denmark.\\
$^{4}$ Max Planck Institute for Intelligent Systems (MPI-IS), Tübingen, Germany.\\
Emails: \tt\small hadi.beik-mohammadi@de.bosch.com, sohau@dtu.dk, gear@tuebingen.mpg.de\\ gerhard.neumann@kit.edu, leonel.rozo@de.bosch.com}
}

\maketitle

\begin{abstract}
*For robots to work alongside humans and perform in unstructured environments, they must learn new motion skills and adapt them to unseen situations on the fly. 
This demands learning models that capture relevant motion patterns, while offering enough flexibility to adapt the encoded skills to new requirements, such as dynamic obstacle avoidance. 
We introduce a Riemannian manifold perspective on this problem, and propose to learn a Riemannian manifold from human demonstrations on which geodesics are natural motion skills. 
We realize this with a variational autoencoder (VAE) over the space of position and orientations of the robot end-effector.
Geodesic motion skills let a robot plan movements from and to arbitrary points on the data manifold. They also provide a straightforward method to avoid obstacles by redefining the ambient metric in an online fashion. 
Moreover, geodesics naturally exploit the manifold resulting from multiple--mode tasks to design motions that were not explicitly demonstrated previously.
We test our learning framework using a $7$-DoF robotic manipulator, where the robot satisfactorily learns and reproduces realistic skills featuring elaborated motion patterns, avoids previously--unseen obstacles, and generates novel movements in multiple-mode settings.  

\end{abstract}

\IEEEpeerreviewmaketitle

\section{Introduction}

Robot motion generation has been actively investigated during the last decades, where motion planners and movement primitives have led to significant advances. 
When a robot moves in obstacle-free environments, the motion generation problem can be easily solved by off-the-shelf motion planners~\cite{Elbanhawi14:MotionPlanning}. 
However, the problem is significantly more involved in unstructured environments when (static and dynamic) obstacles occupy the robot workspace~\cite{Mohanan18:DynamicMotPlan}. 
Moreover, if the robot motion depends on variable targets, or requires to consider multiple-solution tasks, the motion generation problem exacerbates.
Some of the aforementioned problems have been recently addressed from a learning-from-demonstration (LfD) perspective, where a skill model is learned by extracting the relevant motion patterns from human examples~\cite{Osa2018:Imitation}. 

LfD approaches are advantageous as they do not necessarily require a model of the environment, and can easily adapt to variable targets on the fly~\cite{Osa2018:Imitation}.
Three main lines of work stand out in the LfD field, namely, \emph{(1)} dynamical-system-based approaches~\cite{Ijspeert13:dmp} which focus on capturing motion dynamics, \emph{(2)} probabilistic methods~\cite{calinon2016:tutorial,paraschos2018:ProMP,Huang19:KMP} which exploit data variability and model uncertainty, and more recently, \emph{(3)} neural networks~\cite{Yunus2019:CNMPs,Bahl20:NDPs} which address generalization problems. Despite their significant contributions (see Section~\ref{sec:relatedwork}), several challenges are still open: encoding and reproduction of full-pose end-effector movements, skill adaptation to unseen or dynamic obstacles, handling multiple-solution (a.k.a. multiple-mode) tasks, generalization to unseen situations, among others.

\begin{figure}[!t]
	\centering
	\includegraphics[width=0.98\linewidth, ,bb=0 0 1325 995]{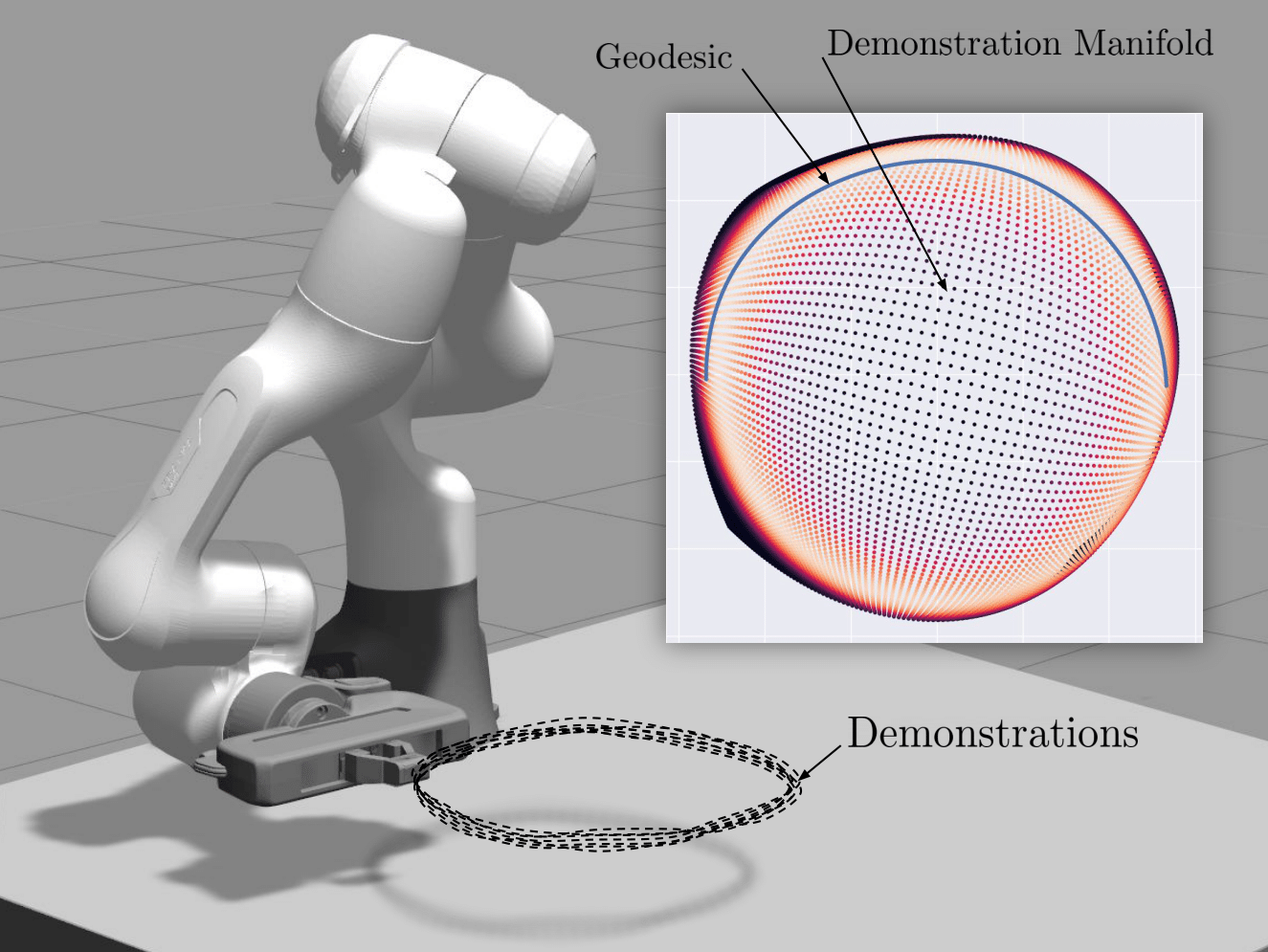}
	\caption{From demonstrations we learn a variational autoencoder that spans a random Riemannian manifold. Geodesics on this manifold are viewed as motion skills.}
	\label{fig:Teaser}
\end{figure}

\textbf{In this paper} we provide an LfD approach that addresses several of these problems, through a Riemannian perspective for learning robot motions from demonstrations. 
Unlike previous works~\cite{Havoutis:MotionPlanningManifold13,Li:TaskManifoldConstrainedManip18}, where skill manifolds are built from locally smooth manifold learning~\cite{Dollar07:LSML}, we leverage a Riemannian formulation. 
Specifically, we develop a variational autoencoder (VAE) that learns a Riemannian submanifold of $\mathbb{R}^3 \times \mathcal{S}^3$ from human demonstrations.
We exploit geodesics (i.e.\ shortest paths) on this learned manifold as the robot motion generation mechanism, from which full-pose end-effector trajectories are generated.
These geodesics reproduce motions starting from arbitrary initial points on the manifold, and
they can adapt to avoid dynamic obstacles in the robot environment by redefining the ambient metric (see Fig.~\ref{fig:Teaser} for illustration).

Our approach can learn manifolds encoding multiple-solution tasks, from which novel (unobserved) robot motions may naturally arise when computing geodesics.

We illustrate our approach with a simple example, and we test our method in robotic experiments using a $7$-DoF robotic manipulator. 
The experiments show how our approach learns and reproduces realistic robot skills featuring complex motion patterns and obstacle avoidance. 
Moreover, we demonstrate how our approach can discover new solutions for unseen task setups in a multiple--mode setting.    

In summary, we contribute a new view on motion skill learning that navigates along geodesics on a manifold learned via a novel VAE over position-orientation space. We show how this allows the robot to generate useful movements beyond the demonstration set, while avoiding dynamic obstacles that were not part of the learning procedure.

\section{Background and related work}
\label{sec:relatedwork}

We first briefly review some relevant work on learning from demonstrations, variational autoencoders (VAEs), and Riemannian geometry. We also introduce some recent connections between VAEs and Riemannian geometry, which is the backbone of our work. 

\subsection{Learning robot motion from demonstrations} 
Learning from demonstrations (LfD) provides a framework for a robot to quickly learn tasks by \emph{observing} several demonstrations provided by a (human) expert~\cite{Osa2018:Imitation}. The demonstrations are then used to learn a model of the task (e.g., a movement primitive, a reward, or plan) which is then used to synthesize new robot motions~\cite{Ravichandar20:LfD}. In particular, movement primitives (MPs) describe complex motion skills, and represent an alternative solution to classic motion planners~\cite{Elbanhawi14:MotionPlanning} for generating robot motions. We here exploit LfD to learn a Riemannian skill manifold, which we later employ to drive the robot motion generation.  

LfD approaches can be categorized into: \emph{(1)} dynamical-system-based approaches~\cite{Ijspeert13:dmp} which capture the demonstrated motion dynamics~\cite{DMP}, \emph{(2)} probabilistic methods~\cite{calinon2016:tutorial,paraschos2018:ProMP,Huang19:KMP} that take advantage of data variability and model uncertainty, and \emph{(3)} neural networks~\cite{Yunus2019:CNMPs,Bahl20:NDPs} aimed at generalization problems. 
Our method leverages a neural network (VAE) to learn a Riemannian metric that incorporates the network uncertainty. This metric allows us to generate motions that resemble the demonstrations. Unlike the aforementioned approaches, our method allows for online obstacle avoidance by rescaling the learned metric. Although obstacle avoidance might still be possible by defining via-points in methods like~\cite{paraschos2018:ProMP,Yunus2019:CNMPs,Huang19:KMP}, this problem was not explicitly considered in any of them.     

Finally, human demonstrations may show different solution trajectories for the same task~\cite{Rozo11:multipleSol,Yunus2019:CNMPs}, which is often tackled through hierarchical approaches~\cite{Konidaris12:SkillTrees, Ewertonetal2015:multcollab}. In this context, our method permits to encode multiple-solution tasks into the learned Riemannian manifold, which is exploited to not only reproduce the demonstrated solutions but also to come up with a hybrid solution built on a synergy of a subset of the demonstrations. These novel solutions naturally arise from our geodesic motion generator. Note that previous learning frameworks generate robot motions that are restricted to the provided solutions for the task at hand.

\subsection{Variational autoencoders (VAEs)}
A variational autoencoder (VAE)~\cite{kingma:autoencoding} is a generative model that captures the data density $\Prob(\x)$ through a latent variable $\z$ that generally has a significantly lower dimension than $\x$. In the interest of simplicity, we consider Gaussian VAEs corresponding to the generative model
\begin{align}
    \Prob(\z) &= \mathcal{N}\left(\z | \bm{0}, \I_d\right), & \z \in \latent ; \\
    \Prob_{\bm{\theta}}(\x|\z) &= \mathcal{N}\left(\z|\mu_{\bm{\theta}}(\z), \I_D \sigma_{\bm{\theta}}^2(\z)\right), & \x \in \ambient.
    \label{eq:vae_gen}
\end{align}
Here $\mu_{\bm{\theta}} : \latent \rightarrow \ambient$ and $\sigma_{\bm{\theta}} : \latent \rightarrow \R^D_+$ are deep neural networks with parameters $\bm{\theta}$ that estimate the mean and the variance of the posterior distribution $\Prob_{\bm{\theta}}(\x|\z)$.

When the generative process is nonlinear, exact inference becomes intractable, and VAEs apply a variational approximation of the evidence (marginal likelihood). The corresponding evidence lower bound (ELBO) is then
\begin{align}
\begin{split}
    \mathcal{L}_{ELBO}
      &= \mathbb{E}_{q_{\bm{\phi}}(\z|\x)}\left[\log(\Prob_{\bm{\theta}}(\x|\z))\right] \\
      &- \mathrm{KL}\left(q_{\bm{\phi}}(\z|\x)||\Prob(\z)\right) ,
    \label{eq:elbo}
\end{split}
\end{align}
where $q_{\bm{\phi}}(\z|\x) = \mathcal{N}(\x | \mu_{\bm{\phi}}(\x), \I_d\sigma^2_{\bm{\phi}}(\x))$ approximates the posterior distribution $p(\z | \x)$ by two deep neural networks $\mu_{\bm{\phi}}(\x) : \ambient \rightarrow \latent$ and $\sigma_{\bm{\phi}}(\x)): \ambient \rightarrow \R^d_+$. 
This approximate posterior is often denoted the \emph{inference} or \emph{encoder} distribution, while the generative process $p_{\bm{\theta}}(\x|\z)$ is known as the \emph{decoder}. As mentioned previously, we use VAEs to learn a robot skill.

\subsection{Riemannian geometry}
Riemannian manifolds can be intuitively seen as curved $d$-dimensional surfaces that are described by smoothly-varying positive-definite inner products, characterized by the Riemannian metric $\Metric$~\cite{Lee18Riemann}. These manifolds locally resemble a Euclidean space $\mathbb{R}^d$, and have a globally defined differential structure. For our purposes, it suffices to consider manifolds as defined by a mapping function
\begin{align}
  f: \latent \rightarrow \ambient,
  \label{eq:mapping}
\end{align}
where both $\latent$ and $\ambient$ are open subsets of Euclidean spaces with $\dim{\latent} < \dim{\ambient}$. We then say that $\Manifold = f(\latent)$ is a manifold immersed in the ambient space $\ambient$.

Given a curve $\curve: [0, 1] \rightarrow \latent$, we can measure its length on the manifold as
\begin{align}
  \Length_{\curve}  &= \int_0^1 \| \partial_t f(\curve(t)) \| \mathrm{d}t .
  \label{eq:length}
\end{align}
By applying the chain-rule, we see that this can be equivalently expressed as
\begin{align}
  \Length_{\curve}  &= \int_0^1 \sqrt{\dot{\curve}(t)^{\trsp} \Metric(c(t)) \dot{\curve}(t)} \mathrm{d}t.
  \label{eq:length_chain_rule}
\end{align}
Here $\dot{\curve}_t = \partial_t \curve_t$ is the curve derivative and where we have introduced the \emph{Riemannian metric}
\begin{equation}
    \Metric(\z) = \Jac_f(\z)^{\trsp} \Jac_f(\z),
    \label{eq:RiemMetric}
\end{equation}
where $\Jac_f$ is the Jacobian of $f$ that we evaluate at $\z \in \latent$. We may think of the metric as forming a local inner product in $\latent$ that inform us how to measure lengths locally. This construction relies on the Euclidean metric of $\ambient$; we will later extend this to also form a Riemannian metric.
Having defined a notion of \emph{curve length} \eqref{eq:length}, we can trivially define shortest paths, or \emph{geodesics}, as curves of minimal length. Geodesics are the generalization of straight lines on the Euclidean space to Riemannian manifolds. They will serve as our motion generation mechanism as explained in Section~\ref{sec:geodesic_motion}. Note that geodesics have been recently used as solutions of trajectory optimizers for quadrotors control~\cite{ScannellTrajectory2021}.   

\begin{SCfigure*}
	\includegraphics[width=0.7\textwidth]{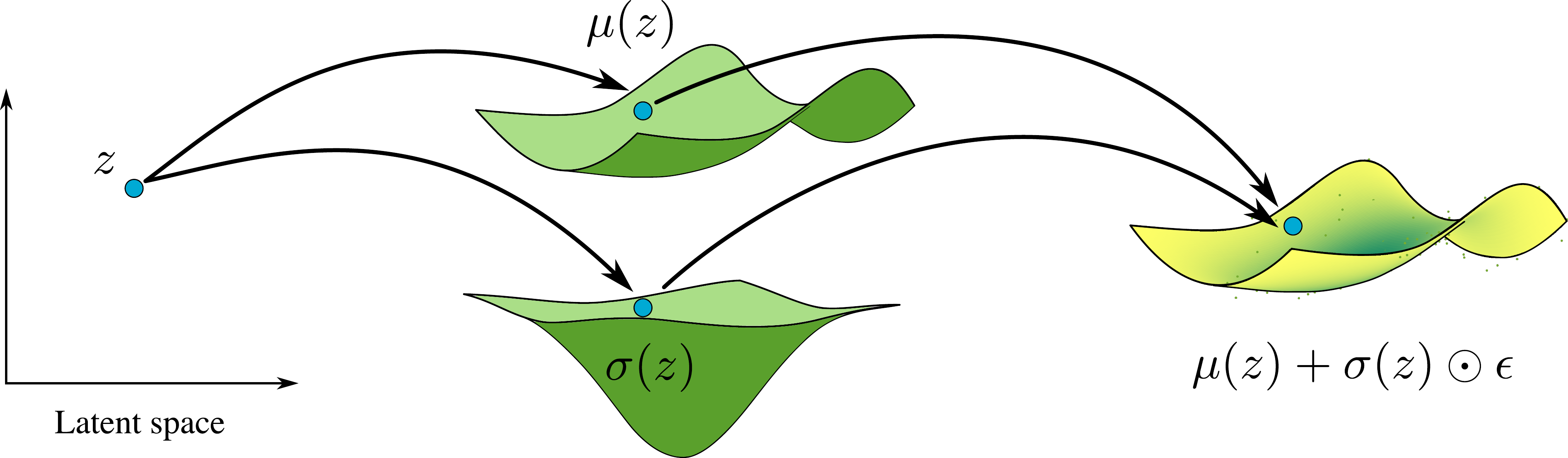}
	\caption{In a Gaussian VAE, samples are generated by a random projection of the manifold jointly spanned by $\mu$ and $\sigma$.}
	\label{fig:Noisy_manifold}
\end{SCfigure*}

\subsection{Variational autoencoders as Riemannian manifolds}\label{sec:vae_manifold}
To make the link between VAEs and Riemannian geometry \cite{Arvanitidis:LatentSO}, we may write the generative process of a VAE \eqref{eq:vae_gen} as
\begin{equation}
    f_{\bm{\theta}}(\z) = \mu_{\bm{\theta}}(\z) + \text{diag}(\epsilon)\sigma_{\bm{\theta}}(\z), \quad \epsilon \sim \mathcal{N}(\bm{0}, \I_D) .
    \label{eq:StochasticF}
\end{equation}
This is also known as the \emph{reparametrization trick}~\cite{kingma:autoencoding} and is illustrated in Fig.~\ref{fig:Noisy_manifold}. We see that this is similar to the mapping function~\eqref{eq:mapping} that defined a manifold in the previous section. The difference being that now $f_{\bm{\theta}}$ is stochastic. We can rewrite this stochastic mapping as~\cite{eklund:arxiv:2019}
\begin{align}
  f_{\bm{\theta}}(\z) &= \begin{pmatrix} \I_D, & \text{diag}(\epsilon) \end{pmatrix} \begin{pmatrix} \mu_{\bm{\theta}}(\z) \\ \sigma_{\bm{\theta}}(\z) \end{pmatrix}
     = \bm{P}\;g(\z) ,
\end{align}
where $\bm{P}$ is a random matrix, and $g(\z)$ is the concatenation of $\mu_{\bm{\theta}}(\z)$ and $\sigma_{\bm{\theta}}(\z)$. In this notation, we can view the VAE as a random projection of a deterministic manifold spanned by $g$, and the metric under this mapping is
\begin{equation}
    \bar{\Metric}(\z) = \Jac_{\mu_{\bm{\theta}}}(\z)^{\trsp} \Jac_{\mu_{\bm{\theta}}}(\z) + \Jac_{\sigma_{\bm{\theta}}}(\z)^{\trsp} \Jac_{\sigma_{\bm{\theta}}}(\z).
    \label{eq:VAE_metric}
\end{equation}
Geodesics under this metric have been shown to be faithful to the data used for training the VAE~\cite{Arvanitidis:LatentSO}. \citet{Hauberg:OnlyBS} argues that this is due to the contribution from $\sigma$ to the metric and that disregarding this term gives an almost flat manifold geometry.

As mentioned, the definition of curve length relies on the Euclidean metric of the ambient space $\ambient$, but this is not a strict requirement.
\citet{arvanitidis:arxiv:2020} argue that there is value in giving the ambient space a manually defined \emph{Riemannian} metric and including that into the definition of curve length. Curve length is then defined as
\begin{align}
    \Length_{\curve}  &= \int_0^1 \sqrt{\dot{\curve}(t)^{\trsp} \Jac_{f_{\bm{\theta}}}(c(t))^{\trsp} \Metric_{\ambient}(f_{\bm{\theta}}(c(t))) \Jac_{f_{\bm{\theta}}}(c(t)) \dot{\curve}(t)} \mathrm{d}t,
\end{align}
where $\Metric_\ambient$ is the ambient space metric, which can now vary smoothly across $\ambient$.
The corresponding Riemannian metric of $\latent$ is then
\begin{align}
    \bar{\Metric}(\z) &= \Jac_{\mu_{\bm{\theta}}}(\z)^{\trsp} \Metric_\ambient(\mu_{\bm{\theta}}(\z)) \Jac_{\mu_{\bm{\theta}}}(\z) \nonumber\\
    &+ \Jac_{\sigma_{\bm{\theta}}}(\z)^{\trsp} \Metric_\ambient(\mu_{\bm{\theta}}(\z)) \Jac_{\sigma_{\bm{\theta}}}(\z) .
    \label{eq:ambient_VAE_metric}
\end{align}
With this construction, it is straightforward to push geodesics away from certain regions of $\ambient$ by increasing $\Metric_{\ambient}$ there.

Note that geodesics do generally not follow a closed-form expression in these models, and numerical approximations are in order. This can be done by direct minimization of curve length~\cite{Shao:TheRiemannianGeometry, kalatzis:icml:2020}, $A^*$ search~\cite{Chen2019:FastApproximateGeodesics}, integration of the associated ODE~\cite{arvanitidis:aistats:2019}, or various heuristics~\cite{chen:MetricsforDeep}.


\section{Riemannian manifold learning on $\mathbb{R}^3 \times \mathcal{S}^3$}
\label{sec:VAEr3s3}
  Learning complex robot motion skills requires models that have enough capacity to learn and synthesize the relevant patterns of a motion while being flexible enough to adapt to new conditions. In this section, we describe how we tackle this problem by bringing a Riemannian manifold perspective to the robot learning problem. First, we explain how we exploit VAEs to access a low-dimensional learned manifold of the motion data where an ambient-space Riemannian metric is learned. This metric will be later used to generate robot motion trajectories via geodesics, as detailed in Section~\ref{sec:geodesic_motion}.
  In order to learn elaborated motion skills, which may display complex position and orientation trajectories, we represent the robot state
  as the full pose of the robot end-effector, i.e.\ its position $\x \in \mathbb{R}^3$ and orientation $\q \in \mathcal{S}^3$. We then seek
  a VAE that models a joint density of this state. We retain the usual Gaussian prior $p(\z) = \mathcal{N}(\z | \bm{0}, \I_d)$, but alter
  the generative distribution $p_{\bm{\theta}, \bm{\psi}}(\x, \q | \z)$ to suit our needs. We will assume that position and orientation are conditionally
  independent,
  \begin{align}
    p_{\bm{\theta}, \bm{\psi}}(\x, \q | \z) = p_{\bm{\theta}}(\x | \z) p_{\bm{\psi}}(\q | \z) ,
  \end{align}
  such that all correlations between the two must be captured by the latent variable $\z$.
  
\subsection{Position encoding on $\mathbb{R}^3$}

    To model the conditional distribution of end-effector positions $\x$, we opt for simplicity and choose this to be
    Gaussian,
    \begin{align}
      p_{\bm{\theta}}(\x | \z) &= \mathcal{N}(\x | \mu_{\bm{\theta}}(\z), \I_3 \sigma^2_{\bm{\theta}}(\z)),
    \end{align}
    where $\mu_{\bm{\theta}}$ and $\sigma_{\bm{\theta}}$ are neural networks parametrized by $\bm{\theta}$. One could argue that
    $p_{\bm{\theta}}(\x | \z)$ should have zero probability mass outside the workspace of the robot, but we disregard
    such concerns as $\sigma_{\bm{\theta}}^2$ tends to take small values due to limited data noise. This implies that
    only a negligible fraction of the probability mass falls outside the robot workspace.

  \subsection{Orientation encoding on $\mathcal{S}^3$}
    Complex robot motions often involve elaborated orientation trajectories which require a suitable representation for motion learning and control. There exist several orientation representation such rotation matrices, Euler angles, and unit quaternions. 
    Euler angles and rotation matrices are commonly used for reasons of simplicity. Unfortunately, Euler angles suffer from gimbal lock \cite{Hemingway2018:gimballock} which makes them an inadequate representation of orientation in robotics, and rotation matrices are a redundant representation requiring a high number of parameters. Unit quaternions are a convenient way to represent an orientation since they are compact, not redundant, and prevent gimbal lock. Also, they provide strong stability guarantees in closed--loop orientation control~\cite{Camarillo08:quaternions}, they have been recently exploited in robot skills learning~\cite{Rozo2020:SkillsSeq}, and for data-efficient robot control tuning~\cite{Jaquier2019:GaBO} under a Riemannian-manifold formulation.
    
    We choose to represent orientations $\q$ as a unit quaternion, such that $\q \in \mathcal{S}^3$ with the additional antipodal identification that $\q$ and $-\q$ correspond to the same orientation. Formally, a unit quaternion $\q$ lying on the surface of a $3$-sphere $\mathcal{S}^3$ can be represented using a $4$-dimensional unit vector $\q = [q_w, q_x, q_y, q_z] \in \mathcal{S}^3$, where the scalar $q_w$ and vector $(q_x, q_y, q_z)$ represent the real and imaginary parts of the quaternion, respectively. 
    To cope with antipodality, one could opt to model $\q$ as a point in a projective space, but for reasons of simplicity we let $\q$ live on the unit sphere $\mathcal{S}^3$. We then choose a generative distribution $p_{\bm{\psi}}(\q | \z)$ such that $p_{\bm{\psi}}(\q | \z) = p_{\bm{\psi}}(-\q | \z)$.

    To construct a suitable distribution $p_{\bm{\psi}}(\q | \z)$ over the unit sphere, we turn to the von Mises-Fischer (vMF) distribution, which is merely an isotropic Gaussian constrained to lie on the unit sphere \cite{Sra18:DirectionalStats}. This distribution is described by a mean direction $\bm{\mu}$ with $\left \| \bm{\mu} \right \| = 1$, and a concentration parameter $\kappa \ge 0$.
    Its density function takes the form
    \begin{align}
      \vMF(\q | \bm{\mu}, \kappa) = C_{D}(\kappa) \exp\left({\kappa\bm{\mu}^{\trsp} \q}\right) ,
      \qquad \| \bm{\mu} \| = 1,
      \label{eq:vmf_density}
    \end{align}
    where $C_{D}$ is the normalization constant
    \begin{align}
    C_{D}({\kappa}) = \frac{\kappa^{\frac{D}{2}-1}}{(2\pi)^{\frac{D}{2}} \mathit{I}_{\frac{D}{2}-1}(\kappa)} ,
    \end{align}
    with $\mathit{I}_{\frac{D}{2}-1}(\kappa)$ being the modified Bessel function of the first kind. Like the Gaussian, from which the distribution was constructed, the von Mises-Fischer distribution is unimodal. To build a distribution that is antipodal symmetric, i.e.\ $p_{\bm{\psi}}(\q | \z) = p_{\bm{\psi}}(-\q | \z)$, we simply form a mixture of antipodal von Mises-Fischer distributions \cite{hauberg:tpami:grassmann},
    \begin{align}
      p_{\bm{\psi}}(\q | \z) &= \frac{1}{2} \vMF(\q | \bm{\mu}_{\bm{\psi}}(\z), \kappa_{\bm{\psi}}(\z)) \nonumber\\
      &+ \frac{1}{2} \vMF(\q | -\bm{\mu}_{\bm{\psi}}(\z), \kappa_{\bm{\psi}}(\z)),
    \end{align}
    where $\bm{\mu}$ and $\kappa$ are parametrized as neural networks. This mixture model is conceptually similar to a Bingham distribution~\cite{Sra18:DirectionalStats}, but is easier to implement numerically.

  \subsection{Variational inference}
    Our VAE model can be trained by maximizing the conventional evidence lower bound (ELBO) \eqref{eq:elbo}, which now is
    \begin{align}
       \label{eq:final_elbo}
       \Loss_{ELBO} &= \alpha \Loss_\x +  \beta \Loss_\q - \mathrm{KL}\left(q_{\bm{\phi}}(\z|\x)||\Prob(\z)\right),\\
       \Loss_\x &=  \mathbb{E}_{q_{\bm{\phi}}(\z|\x)}\left[\log p_{{\bm{\theta}}}(\x|\z) \right] , \\
       \Loss_\q &=  \mathbb{E}_{q_{\bm{\phi}}(\z|\x)}\left[\log p_{{\bm{\psi}}}(\q|\z) \right] ,
    \end{align}
    where $\x \in \R^3$ and $\q \in \Sph^3$ represent the position and quaternion of the end-effector, respectively.
    To balance the log-likelihood of position and orientation components, $\alpha>0$ and $\beta>0$ are proportionally scaled. Note that due to the antipodal nature of quaternions, raw demonstration data may contain negative or positive values for the same orientation. So, we avoid any pre-processing step of the data by considering two von Mises-Fischer distributions that encode the same orientation at both sides of the hypersphere. Practically, we double the training data, by including $\q_n$ and $-\q_n$ for all observations $\q_n$.

  \subsection{Induced Riemannian metric}
    Our generative process is parametrized by a set of neural networks. Specifically, $\mu_{\bm{\theta}}$ and $\sigma_{\bm{\theta}}$ are position mean and variance neural networks parameterized by $\bm{\theta}$, while $\mu_{\bm{\psi}}$ and $\kappa_{\bm{\psi}}$ are neural networks parameterized by $\bm{\psi}$ that represent the mean and concentration of the quaternion distribution. Following Sec.~\ref{sec:vae_manifold} the Jacobians of these functions govern the induced Riemannian metric as
    \begin{align}
      \label{eq:pos_quat_metric}
      \Metric(\z) &= \Metric_\mu^\x(\z) + \Metric_\sigma^\x(\z) + \Metric_\mu^\q(\z) + \Metric_\kappa^\q(\z)
    \end{align}
    with
    \begin{align*}
      \Metric_\mu^\x(\z) &= \Jac_{\mu_{\bm{\theta}}}(\z)^{\trsp} \Jac_{\mu_{\bm{\theta}}}(\z), \quad \Metric_\sigma^\x(\z) = \Jac_{\sigma_{\bm{\theta}}}(\z)^{\trsp} \Jac_{\sigma_{\bm{\theta}}}(\z),\\
      \Metric_\mu^\q(\z) &= \Jac_{\mu_{\bm{\psi}}}(\z)^{\trsp} \Jac_{\mu_{\bm{\psi}}}(\z), \quad \Metric_\kappa^\q(\z) = \Jac_{\kappa_{\bm{\psi}}}(\z)^{\trsp} \Jac_{\kappa_{\bm{\psi}}}(\z) ,
    \end{align*}
    where $\Jac_{\mu_{\bm{\theta}}}$, $\Jac_{\sigma_{\bm{\theta}}}$, $\Jac_{\mu_{\bm{\psi}}}$, $\Jac_{\kappa_{\bm{\psi}}}$ are the Jacobian of functions representing the position mean and variance, as well as the quaternion mean and concentration, respectively. 

    In practice, we want this metric $\Metric(\z)$ to take large values in regions with little or no data, so that geodesics avoid passing through them. Following \citet{Arvanitidis:LatentSO} we achieve this by using radial basis function (RBF) networks as our variance representation, whose kernels reliably extrapolate over the whole space. Since one of the main differences between Gaussian and von Mises-Fischer distributions lies on the way they represent data dispersion, the RBF network should consider a reciprocal behavior when estimating variance for positions. In summary, the data uncertainty is encoded by the RBF networks representing $\sigma^{-1}_{\bm{\theta}}(\z)$ and $\kappa_{\bm{\psi}}(\z)$, which affect the Riemannian metric through their corresponding Jacobians as in ~\eqref{eq:pos_quat_metric}.

\section{Geodesic Motion Skills}
\label{sec:geodesic_motion}
As mentioned previously, geodesics follow the trend of the data, and they are here exploited to reconstruct motion skills that resemble human demonstrations. Moreover, we explain how new geodesic paths, that avoid obstacles on the fly, can be obtained by a metric scaling process. In particular, we exploit ambient space metrics defined as a function of the obstacles configuration to locally deform the original learned Riemannian metric. Last but not least, our approach can encode multiple-solution skills, from which new hybrid trajectories (not previously shown to the robot) can be synthesized. We elaborate on each of these features in the sequel.

\subsection{Geodesic motion generation}
\label{subsec:geodesicmotion}
Geodesic curves generally follow the trend of the training data, due to the role of uncertainty in the metric. Specifically, Eq.~\eqref{eq:VAE_metric} tells us that geodesics are penalized for crossing through regions where the VAE predictive uncertainty grows.
This implies that if a set of demonstrations follows a circular motion pattern, geodesics starting from arbitrary points on the learned manifold will also generate a circular motion (see Fig. \ref{fig:Teaser}). 
This behavior is due to the way that the metric $\Metric$ is defined, as $\Metric$ is characterized by low values where data uncertainty is low (and vice-versa). Since the geodesics minimize the energy of the curve between two points on $\Manifold$, which is calculated as a function of $\Metric$, they tend to stay on the learned manifold and avoid outside regions. This property makes us suggest that geodesics form a natural motion generation mechanism. 
Note that when using a Euclidean metric (i.e., an identity matrix), geodesics correspond to straight lines. Such geodesics certainly neglect the data manifold geometry.
 
Formally, we compute geodesics on $\Manifold$ by approximating them by cubic splines $\curve \approx {\omega_\lambda}(\z_{c})$,
with $\z_c = \{\z_{c_0}, \ldots, \z_{c_N} \}$, where $\z_{c_n} \in \latent$ is a vector defining a control point of the spline  over the latent space $\latent$. Given $N$ control points, $N-1$ cubic polynomials $\omega_{\lambda_i}$ with coefficients $\lambda_{i,0}$, $\lambda_{i,1}$, $\lambda_{i,2}$, $\lambda_{i,3}$ have to be estimated to minimize its Riemannian length
\begin{equation}
\label{eq:cost_geodesic}
\Loss_{{\omega_\lambda}(\z_{c})} = \int_0^1 \sqrt{\left \langle \dot{\omega}_\lambda(\z_{c}), \Metric(\omega_\lambda(\z_{c}))\dot{\omega}_\lambda(\z_{c}) \right \rangle} \mathrm{d}t .
\end{equation}
Then, the final geodesic $\curve$ computed in $\latent$ is used to generate the robot motion through the mean decoder networks $\mu_{\bm{\theta}}$ and $\mu_{\bm{\psi}}$. The resulting trajectory can be executed on the robot arm to reproduce the required skill.

\begin{figure}
  \centering
  \includegraphics[width=1.0\linewidth]{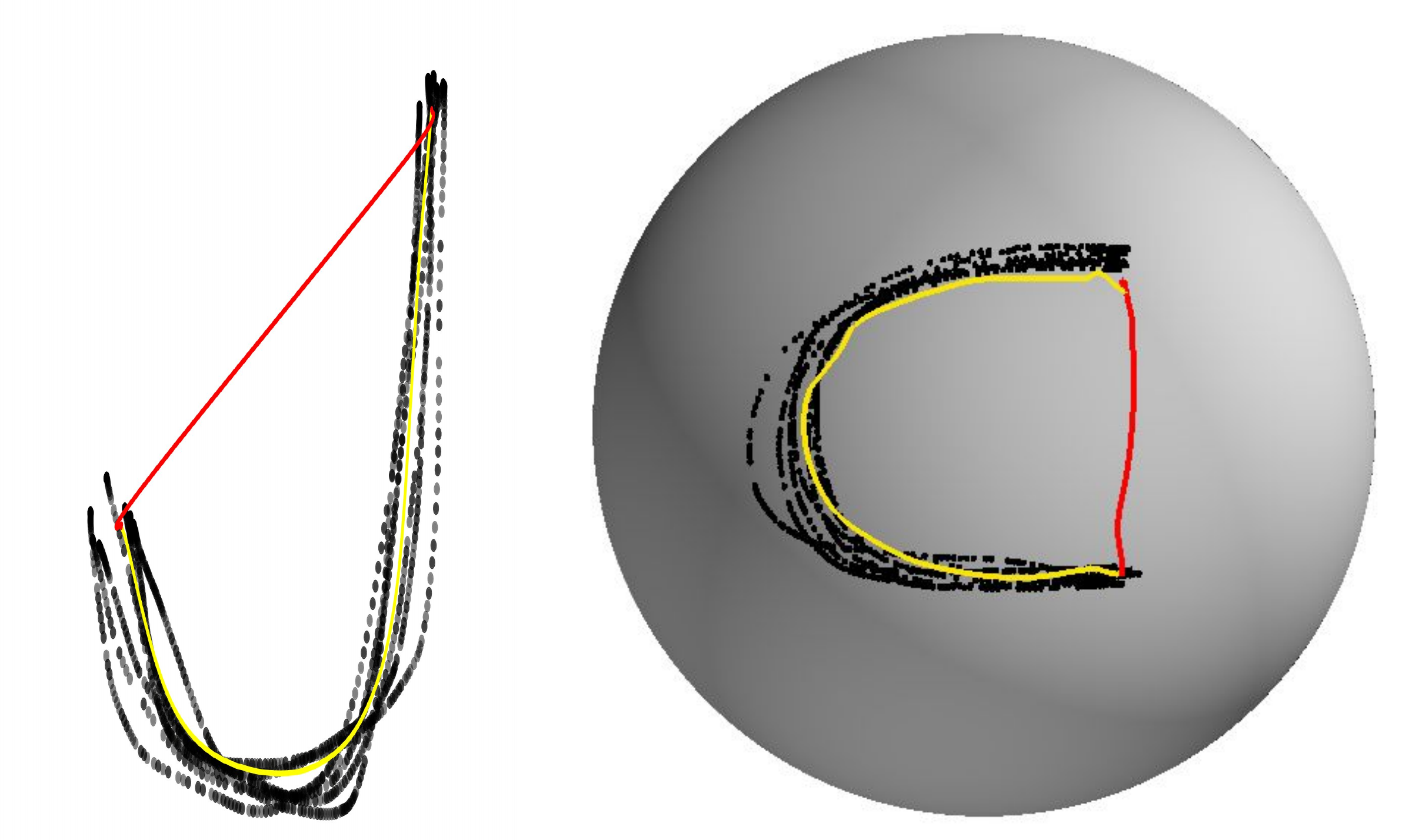}
\caption{As an illustration, we consider synthetic data that belong to $\R^2 \times \Sph^2$. The left panel depicts the $\mathsf{J}$-shaped position data in $\R^2$ and the right panel shows the $\mathsf{C}$-shaped orientation data on $\mathcal{S}^2$. The yellow and red curves show the geodesics computed based on Riemannian and Euclidean metrics, respectively.}
\label{fig:Toy_example_data}
\end{figure}
\begin{figure}[t]
\centering
\begin{subfigure}{\linewidth}
   \includegraphics[width=1\linewidth]{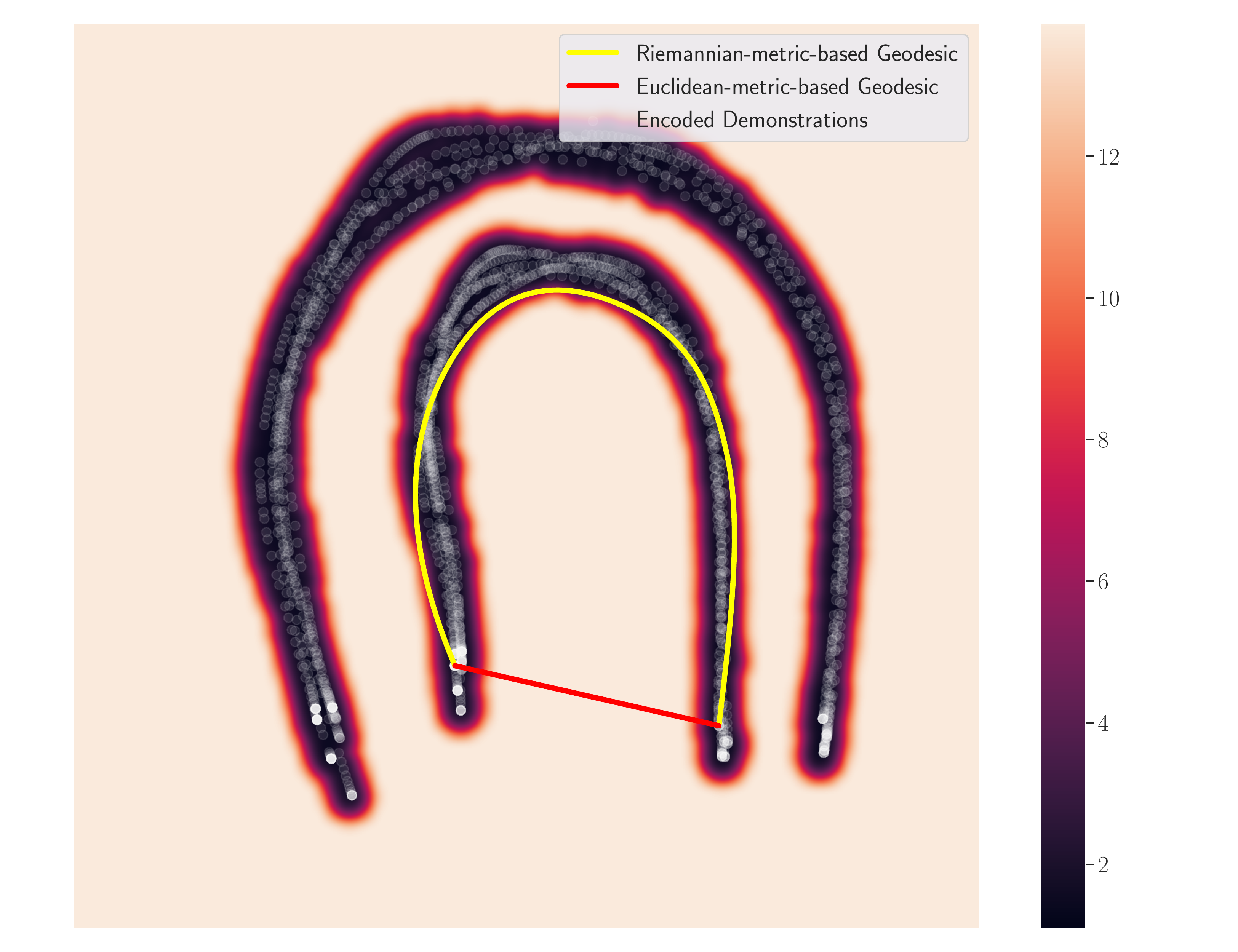}
\end{subfigure}
\begin{subfigure}{\linewidth}
   \includegraphics[width=1\linewidth]{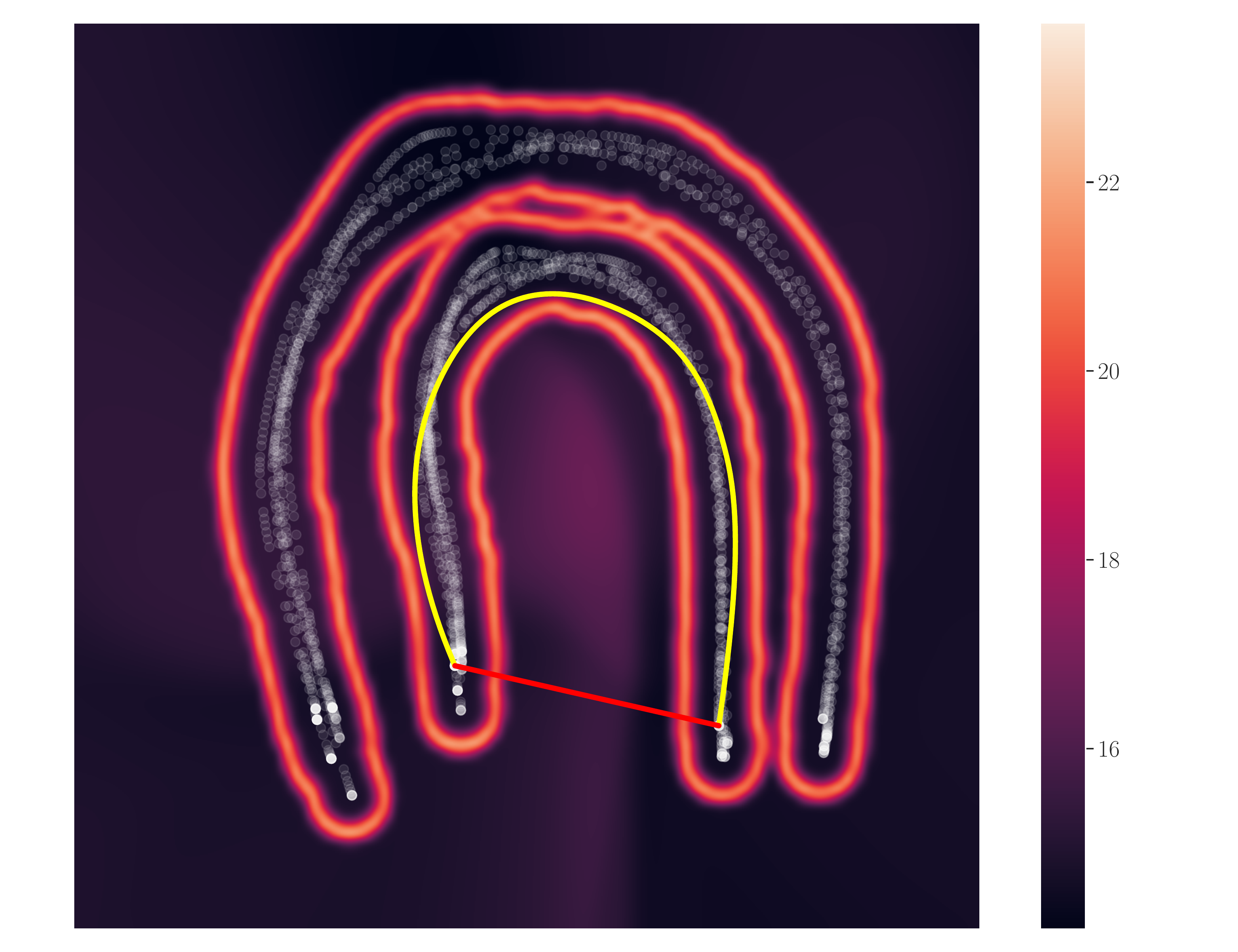}
\end{subfigure}
\caption{ \emph{Top}: the variance measure, \emph{bottom}: magnification factor of the Riemannian manifold learned from trajectories based on $\mathsf{J}$ and $\mathsf{C}$ English alphabet characters defined in $\R^2 \times \mathcal{S}^2$. The semi-transparent white points depict the encoded training set and the yellow curve depicts the geodesic in the latent space. The resulting manifold is composed of two similar clusters due to the antipodal encoding of the quaternions, where each cluster represents one side of the hyper-sphere. The yellow and red curves show the geodesics computed based on Riemannian and Euclidean metrics, respectively.}
\label{fig:Toy_example}
\end{figure}

To illustrate, we consider a simple experiment where the demonstration data at each time point is confined to $\R^2 \times \Sph^2$, i.e.\ only two dimensional positions and orientations are considered. We create synthetic position data that follows a $\mathsf{J}$-shape and orientation data that follows a $\mathsf{C}$-shape projected on the sphere (see Fig.~\ref{fig:Toy_example_data}). We fit our VAE model to this data, and visualize the corresponding latent space in Fig.~\ref{fig:Toy_example}. Here the top panel shows the latent mean embeddings of the training data with a background color corresponding to the predictive uncertainty. We see low uncertainty near the data, and high otherwise. The bottom panel of Fig.~\ref{fig:Toy_example} shows the same embedding but with a background color proportional to $\log\sqrt{\det\Metric}$. This quantity, known as the magnification factor \cite{Bishop1997:magfactor}, will generally take large values in regions where distances are large, implying that geodesics will try to avoid such regions. In the figure, we notice that the magnification factor is generally low, except on the `boundary' of the data manifold, i.e.\ in regions where the predictive variance grows. Consequently, we observe that Riemannian geodesics (yellow curves in the figure) stay within the `boundary' and are hence near the training data. In contrast, Euclidean geodesics (red curves in the figure) fail to stay in the data manifold. Our proposal is to use Riemannian geodesics to generate new motions for the robot.
\subsection{Obstacle avoidance using ambient space metrics}
\label{subsec:obstavoidance}
Often human demonstrations do not include any notion of obstacles in the environment. As a result, obstacle avoidance is usually treated as an independent problem when generating robot motions in unstructured environments. A possible solution to integrate both problems is to provide obstacle--aware demonstrations, where the robot is explicitly taught how to avoid known obstacles. The main drawback here is that the robot is still unable to avoid unseen obstacles on the fly. 

The Riemannian approach provides a natural solution to this problem. The natural metric in latent space \eqref{eq:VAE_metric} measures the length of a latent curve under the Euclidean space of the ambient space $\ambient$. We can easily modify this to take obstacles into account. Intuitively, we can increase the length of curves that intersect obstacles, such that geodesics are less likely to go near the obstacles. Formally, we propose to alter the ambient metric of the end-effector position to be
\begin{align}
  \Metric_\ambient^\x(\x) &= \left( 1 + \eta \exp\left( \frac{-\| \bm{x} - \bm{o} \|^2}{2r^2} \right)\right)\I_3,
  \quad \bm{x} \in \mathbb{R}^3,
\end{align}
where $\eta > 0$ scales the cost, $\bm{o} \in \R^3$ and $r>0$ represent the position and radius of the obstacle, respectively. For the orientation component, we use $\Metric_\ambient^\q(\x) = \I_4$. Under this ambient metric, geodesics will generally avoid the object, though we emphasize this is only a \emph{soft} constraint. This approach is similar in spirit to CHOMP~\cite{Ratliff2009:chomp} except our formulation works along a low-dimensional learned manifold, whose solution is then projected to the task space of the robot.
Under this ambient metric, the associated Riemannian metric of the latent space $\latent$ becomes
\begin{equation}
\label{eq:obstacle_metric}
\Metric(\z) = \Metric_\mu^\x(\z) + \Metric_\sigma^\x(\z) + \Metric_\mu^\q(\z) + \Metric_\kappa^\q(\z) ,
\end{equation}
\begin{align}
\text{with}\quad \Metric_\mu^\x(\z) &= \Jac_{\mu_{\bm{\theta}}}(\z)^{\trsp} \Metric_{\ambient}^\x(\mu_{\bm{\theta}}(\z)) \Jac_{\mu_{\bm{\theta}}}(\z) , \nonumber \\  
\Metric_\sigma^\x(\z) &= \Jac_{\sigma_{\bm{\theta}}}(\z)^{\trsp} \Metric_{\ambient}^\x(\mu_{\bm{\theta}}(\z)) \Jac_{\sigma_{\bm{\theta}}}(\z) , \nonumber \\  
\Metric_\mu^\q(\z) &= \Jac_{\mu_{\bm{\psi}}}(\z)^{\trsp} \Metric_\ambient^\q(\mu_{\bm{\psi}}(\z)) \Jac_{\mu_{\bm{\psi}}}(\z) , \nonumber \\  
\Metric_\kappa^\q(\z) &= \Jac_{\kappa{\bm{\psi}}}(\z)^{\trsp} \Metric_\ambient^\q(\mu_{\bm{\psi}}(\z)) \Jac_{\kappa{\bm{\psi}}}(\z) ,  \nonumber
\end{align}
where $\Metric_{\ambient}^\x$ and $\Metric_{\ambient}^\q$ represent the position and orientation components of the obstacle--avoidance metric $\Metric_\ambient$, respectively. Here we emphasize that as the object changes position, the VAE does not need to be re-trained as the change is only in the ambient metric.

\subsection{Real time motion generation}
  Having phrased motion generation as the computation of geodesics, we evidently need a fast and robust algorithm for computing geodesics.
  As we work with low--dimensional latent spaces, we here propose to simply discretize the latent space on a regular grid and use a graph--based algorithm for computing shortest paths. 

  \begin{figure}
      \centering
      \includegraphics[width=\linewidth]{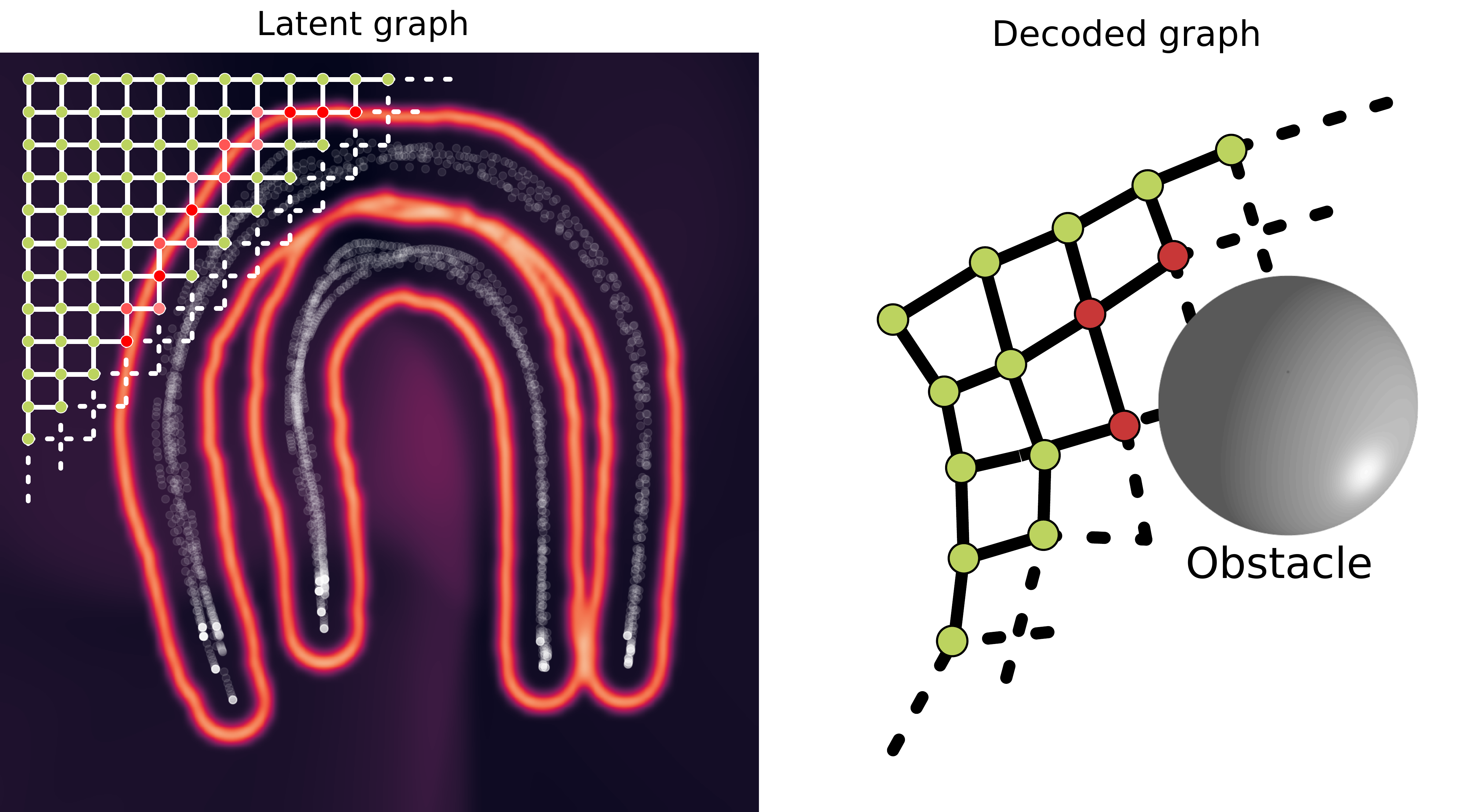}
      \caption{Concept drawing. \emph{Left:} The latent space is discretized to form a grid graph consisting of linearly spaced nodes, with edge weights matching Riemannian distances. \emph{Right:} To efficiently handle obstacles, the graph is decoded, such that obstacles can easily be mapped to latent space.}
      \label{fig:graph}
  \end{figure}
  Specifically, we create a uniform grid over the latent space, and assign a weight to each edge in the graph corresponding to the Riemannian distance between neighboring nodes (see Fig.~\ref{fig:graph}). Geodesics are then found using Dijkstra's algorithm \cite{cormen2009introduction}.
  This algorithm selects a set of graph nodes,
  \begin{align}
      G_\curve = \left \{\g_0, \g_1, \ldots, \g_{N-1}, \g_N \right \}, \quad \g_n \in \R^D, \nonumber
  \end{align}
   where $\g_0$ and $\g_N$ represent the start and the target of the geodesic in the graph, respectively. To select these points, the shortest path on the graph is calculated by minimising the accumulated weight (cost) of each edge connecting two nodes calculated as in \eqref{eq:length_chain_rule}. To ensure a smooth trajectory, we fit a cubic spline  $\omega_\lambda$ to the resulting set $G_c$ by minimizing the mean square error.
  The spline computed in $\latent$ is finally used to generate the robot motion through the mean decoder networks $\mu_{\bm{\theta}}$ and $\mu_{\bm{\psi}}$. The resulting trajectory can be executed on the robot arm to reproduce the required skill.

  One issue with this approach is that dynamic obstacles imply that geodesic distances between nodes may change dynamically. To avoid recomputing all edge weights every time an obstacle moves we do as follows.
  Since the learned manifold does not change, we can keep a decoded version of the latent graph in memory (Fig.~\ref{fig:graph}). This way we need not query the decoders at run-time. We can then find points on the decoded graph that are near obstacles and rescale their weights to take the ambient metric into account. Once the obstacle moves, we can reset the weights of points that are now sufficiently far from the obstacle.
  The center panel of Fig.~\ref{fig:Discrete_manifold} provides an example showing how the metric on the left panel is represented as a discrete graph.

\section{Experiments}
\label{sec:experiments}
We evaluate the performance and capabilities of our method in two different scenarios\footnote{We extensively evaluated our method on simulation settings as the COVID-19 pandemic prohibited access to our robotic labs. We still tested our learning approach with real robot data to validate that our insights apply in realistic scenarios. We plan to run more real experiments whenever possible.}: a simulated pouring task, and a real-world grasping scenario both in $\R^3 \times \Sph^3$. In particular, the pouring task showcases a multiple-solution setting. For the experiments, we discuss the design of each experiment regarding manifold learning and geodesic computation. We also provide a visualization of the learned Riemannian metrics and geodesic representation in the latent space $\latent$. Furthermore, the code is available at: \href{https://sites.google.com/view/geodesicmotion}{https://sites.google.com/view/geodesicmotion}.

\subsection{Setup description}
We consider simulated and real robot demonstrations involving a $7$-DoF Franka Emika Panda robot arm with a two--finger gripper. The demonstrations were recorded using kinesthetic teaching in the real grasping scenario meanwhile simulated pouring dataset was collected using the Franka ROS Interface~\cite{saif:interface} on Gazebo~\cite{saif:gazebo}. In both scenarios, the robot is controlled by an impedance controller. 

We calculate geodesic on a $100 \times 100$ grid graph, and our straightforward Python implementation runs at $100$Hz on ordinary PC hardware. The approach readily runs in real time.

\begin{figure*}[th]
\centering
\begin{subfigure}{.38\textwidth}
  \centering
  \includegraphics[width=\linewidth]{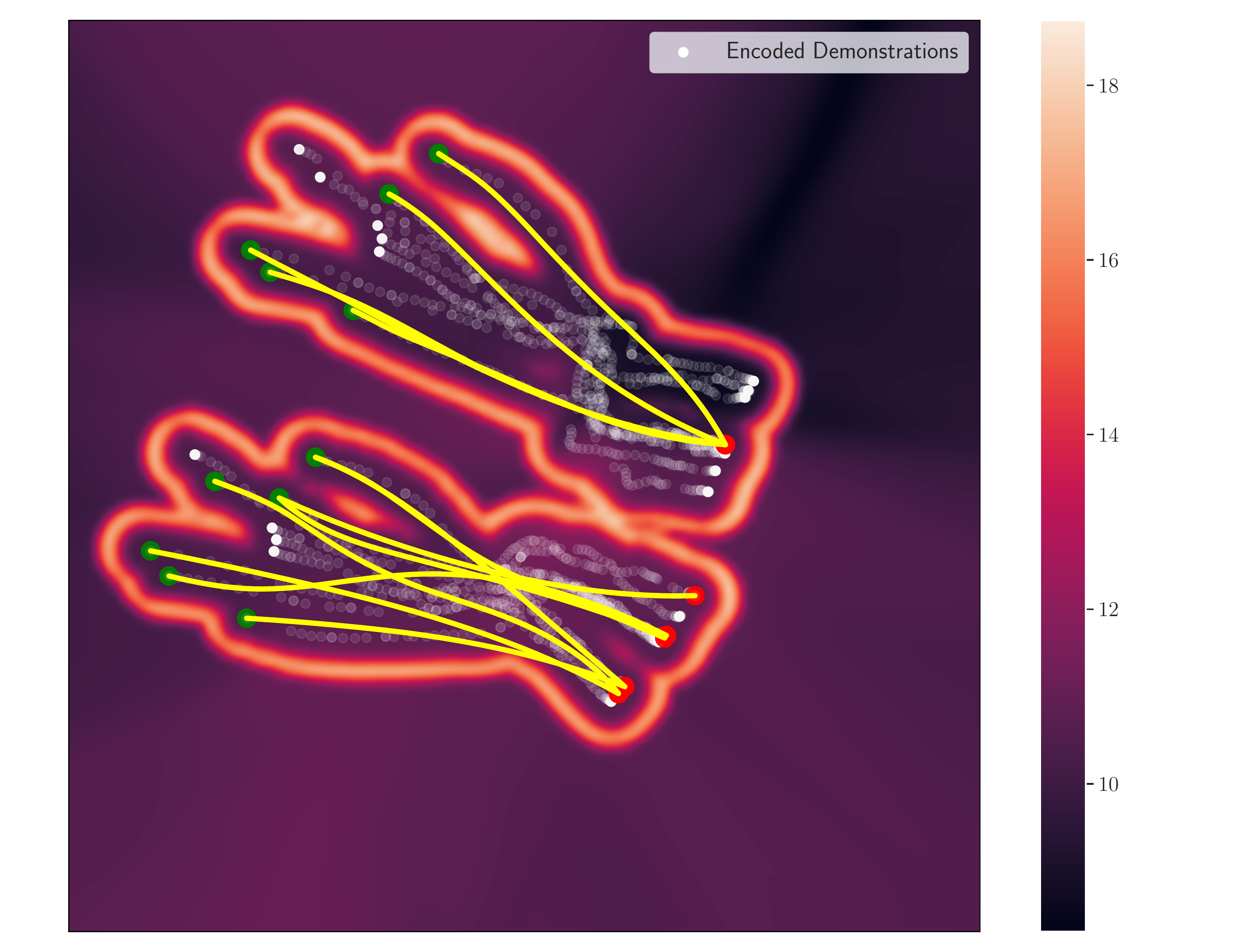}
\end{subfigure}%
\begin{subfigure}{.30\textwidth}
  \centering
 \includegraphics[width=\linewidth]{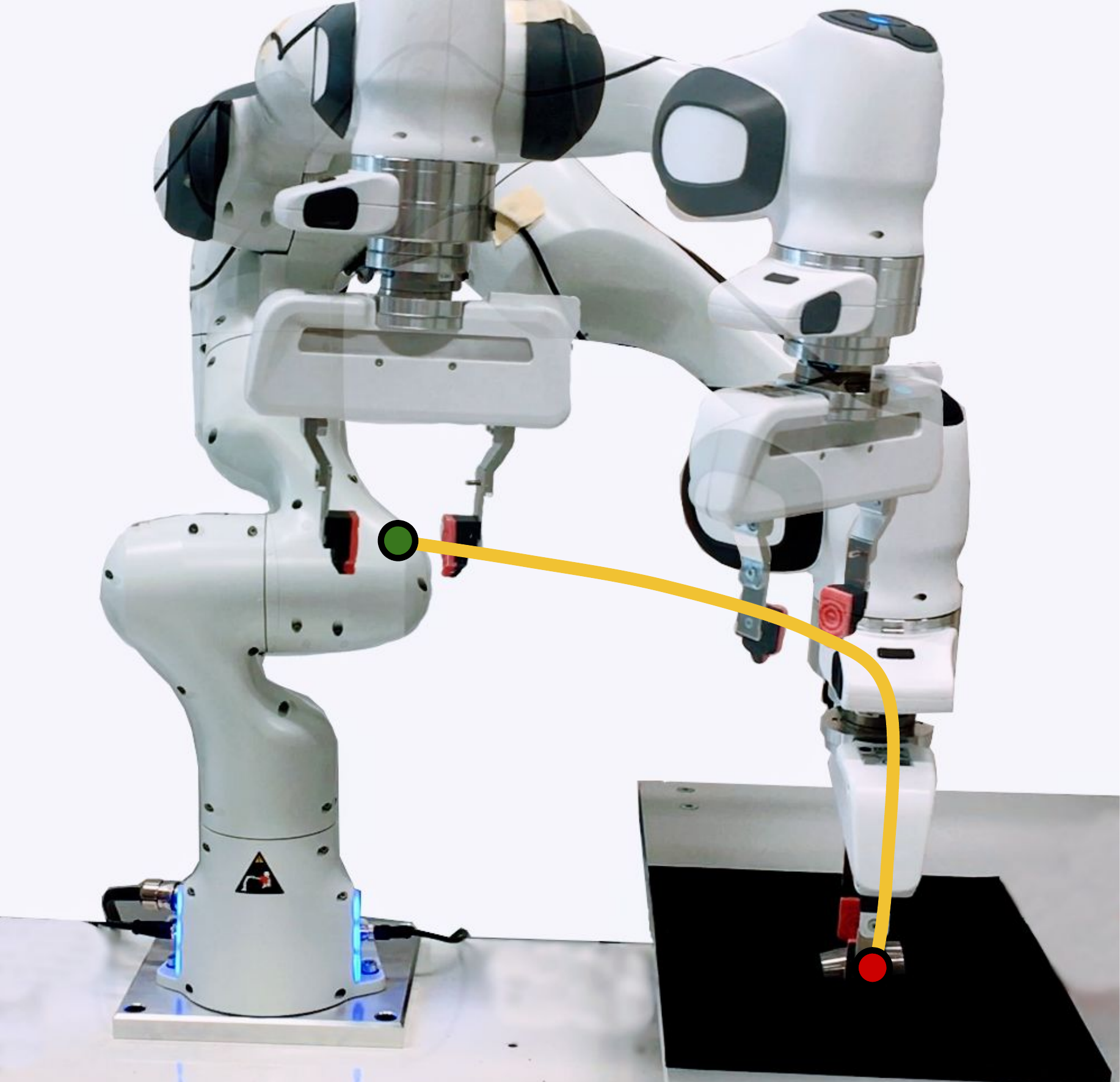}
\end{subfigure}
\begin{subfigure}{.30\textwidth}
  \centering
  \includegraphics[width=\linewidth]{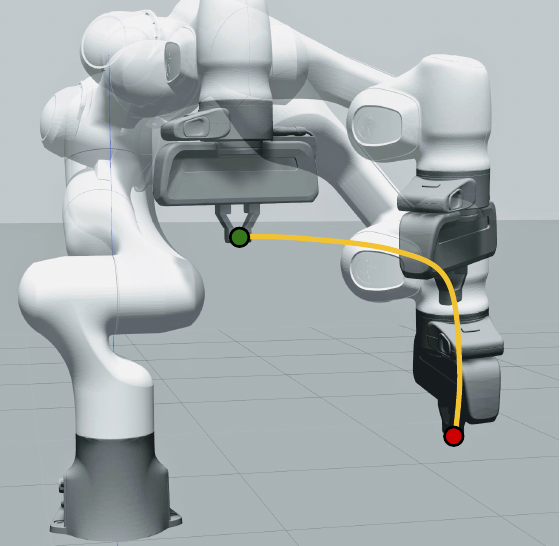}
\end{subfigure}
\caption{\emph{Left}: The yellow curves in the top cluster show geodesics starting from random points and ending up at the same target, and the curves in the bottom cluster connect random points on the manifold. The background depicts magnification factor derived from the learned manifold, and the semi-transparent white points show the encoded training dataset. \emph{Middle}: It illustrates one of the demonstrations on the real robot, where the yellow curve depicts the end-effector trajectory. \emph{Right}: It shows the reconstructed geodesic using the decoder network applied on the simulated robot. The yellow curve depicts the decoded geodesic computed on the learned manifold.}
\label{fig:robot_overlay_sim}
\label{fig:Grasping_geodesic_mf}
\end{figure*}

\subsection{VAE architecture}
Our VAE architecture is implemented using PyTorch~\cite{NEURIPS2019_9015}. The decoder and encoder networks have two hidden layers with $200$ and $100$ neuron units. The same architecture is used for all the experiments. The RBF variance/concentration networks use $500$ kernels calculated by $k$-means over the training dataset \cite{Arvanitidis:LatentSO} and predefined bandwidth. The latent space of the VAE is $2$-dimensional, while the ambient space is $7$-dimensional, corresponding to $\R^3 \times \Sph^3$. 

We employ a single neural network to represent both the position and orientation decoder means, meaning that our final metric is defined as 
\begin{equation}
    \label{eq:obstacle_metric_exp}
    \Metric(\z) = \Metric_{\mu}^{\x,\q}(\z) + \Metric_\sigma^\x(\z) + \Metric_\kappa^\q(\z) , 
\end{equation}
\begin{align}
    \text{with} \quad \Metric_{\mu}^{\x,\q}(\z) &= \Jac_{\mu_{\bm{\theta}}}(\z)^{\trsp} \Metric_{\ambient\Q}(\z)\Jac_{\mu_{\bm{\theta}}}(\z) , \nonumber\\
    \Metric_{\ambient\Q}(\z) &= 
    \begin{bmatrix}
\Metric_\ambient(\mu_{\bm{\theta}}(\z)) & \bm{0} \\  
\bm{0} & \Metric_\Q(\mu_{\bm{\psi}}(\z))
\end{bmatrix} , \nonumber\\
    \Metric_\sigma^\x(\z) &= \Jac_{\sigma_{\bm{\theta}}}(\z)^{\trsp} \Metric_\ambient(\mu^\x_{\bm{\theta}}(\z)) \Jac_{\sigma_{\bm{\theta}}}(\z) , \nonumber\\
    \Metric_\kappa^\q(\z) &= \Jac_{\kappa{\bm{\psi}}}(\z)^{\trsp} \Metric_\Q(\mu_{\bm{\psi}}(\z)) \Jac_{\kappa{\bm{\psi}}}(\z) . \nonumber
\end{align}
where $\Jac_{\mu_{\bm{\theta}}} \in \R^{(D_\ambient + D_\Q) \times d}$ is the Jacobian of the joint decoder mean network (position and quaternion), and $\Jac_{\sigma_{\bm{\theta}}}\in \R^{D_\ambient \times d}$ and $\Jac_{\kappa_{\bm{\psi}}} \in \R^{D_\Q \times d}$ are the Jacobians of the decoder variance and concentration RBF networks. Since the position and quaternion share the same decoder mean network, the output vector is split into two parts, accordingly. The quaternion part of the decoder mean is projected to the $\Sph^3$ to then define the corresponding von Mises-Fischer distribution~\eqref{eq:vmf_density}.

The ELBO parameters $\alpha$ and $\beta$ in \eqref{eq:final_elbo} are found experimentally to guarantee good reconstruction of both position and quaternion data.
It is worth pointing out that we manually provided antipodal quaternions during training, which leads to better latent space structures and reconstruction accuracy.  


\subsection{Real grasping task}
The first set of experiments is based on a dataset collected while a human operator performs kinesthetic demonstrations of a grasping skill. This particular grasping motion includes a $90$\textdegree~rotation when approaching the object for performing a side grasp~\cite{Rozo2020:SkillsSeq}.
The demonstration trajectories start from different end-effector poses, and they reach the same target position with slightly different orientations.

To reproduce the grasping skill, we computed a geodesic in $\latent$ which leads to a continuous trajectory in $\ambient$, that closely reproduces  the rotation pattern observed during demonstrations.

Figure~\ref{fig:Grasping_geodesic_mf} depicts the magnification factor related to the learned manifold. The semi--transparent white points correspond to the latent representation of the training set, and the yellow curves are geodesics between points assigned from the start and endpoints of the demonstrations. The left panel in Fig.~\ref{fig:Grasping_geodesic_mf} shows geodesics in two different scenarios: The ones in the top cluster start from different poses and end up at the same target, and the geodesics in the bottom cluster start and end in different random poses. The target points on the most right side of each cluster represent the same position but due to their slightly different orientation, they are encoded on different latent points. 

\begin{SCfigure*}
  \centering
\centering
\begin{subfigure}{.8\linewidth}
   \includegraphics[width=1\linewidth]{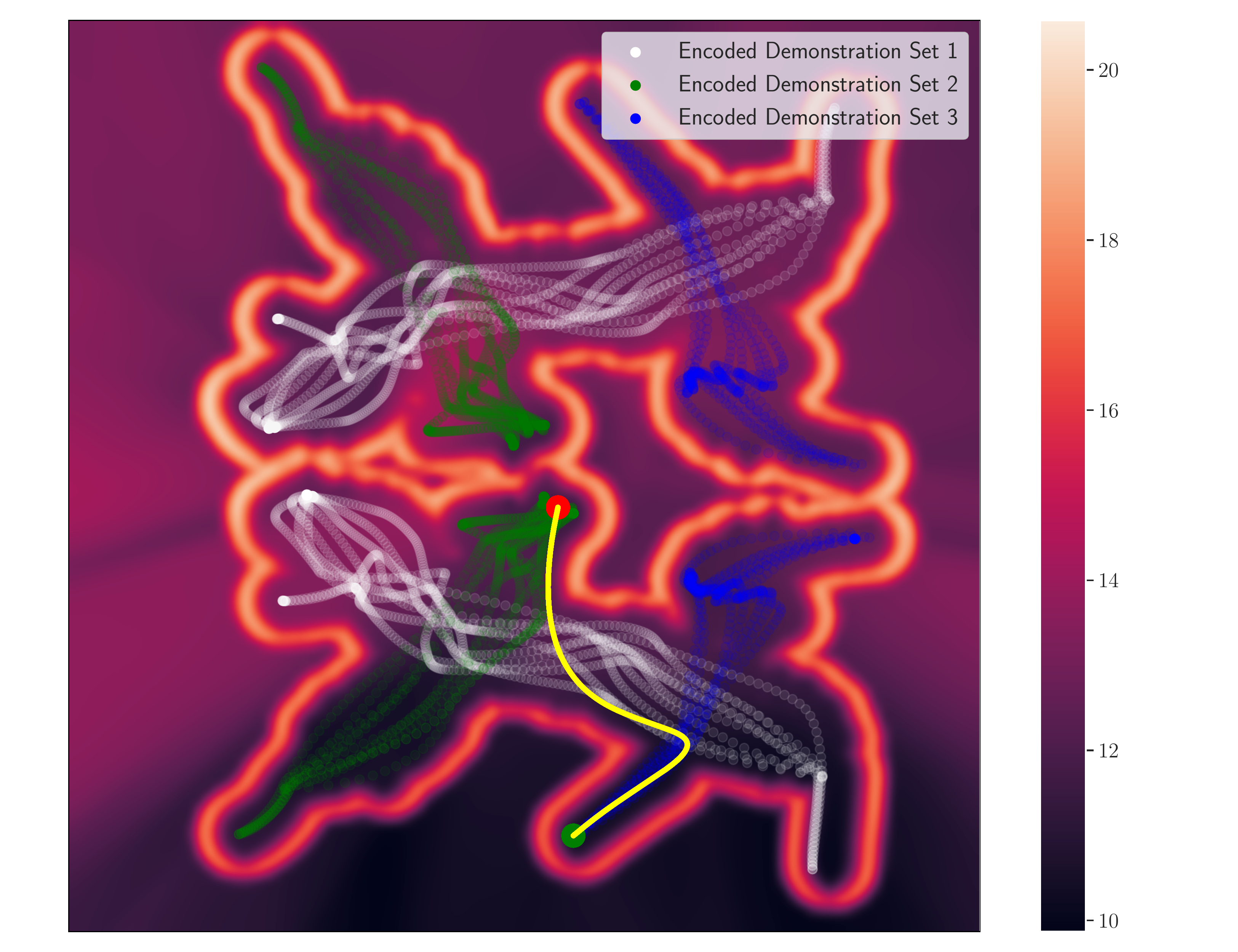}
\end{subfigure}
\begin{subfigure}{.6\linewidth}
   \includegraphics[width=1\linewidth]{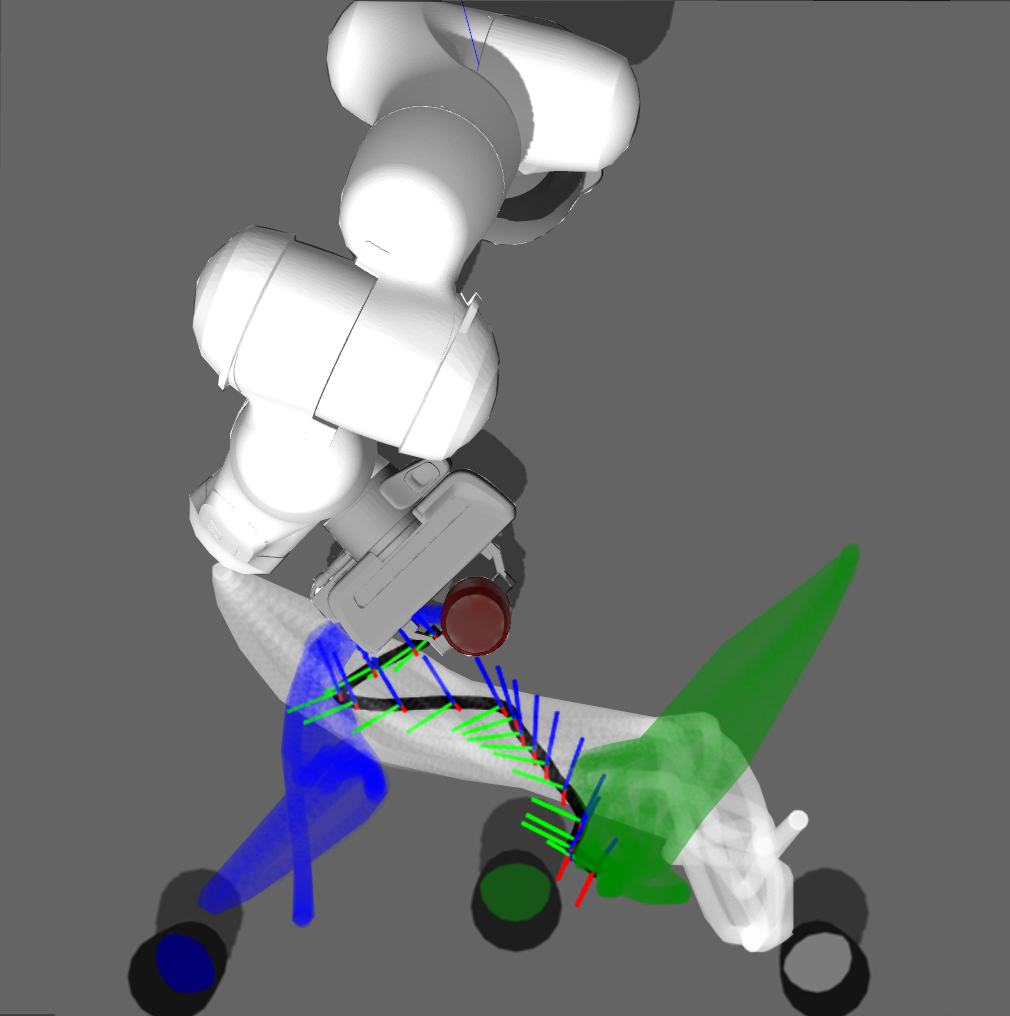}
\end{subfigure}
  \caption{\textit{Left}: The geodesic depicted as the yellow curve takes advantage of the different sets of demonstrations (blue, green and while dots) to generate a hybrid solution which was not explicitly demonstrated to the robot. \textit{Right}: Robot configurations considered in the left plot, from the top perspective in the ambient space. The decoded geodesics are depicted as a black curve and the orientations are visualized by coordinate systems.}
    \label{fig:Multiple_solution}
\end{SCfigure*}

The results show that the method can successfully generate geodesics that respect the geometry of the manifold learned from demonstration. Interestingly, as shown by the magnification factor plot (Fig.~\ref{fig:Grasping_geodesic_mf}-\emph{left}), the resulting manifold is composed of two similar clusters, similarly to the illustrative example of Fig.~\ref{fig:Toy_example}. We observed that this behavior emerged due to the antipodal encoding of the quaternions, where each cluster represents one side of the hyper-sphere. It is worth highlighting that this encoding alleviates any kind of post-processing of raw quaternion data during training or reproduction phases.

Figure~\ref{fig:robot_overlay_sim}-\emph{middle} shows one of the demonstrated trajectories on real robot, and Fig.~\ref{fig:robot_overlay_sim}-\emph{right} displays the reconstructed geodesic using the decoder network and applied on a simulated robot arm. From the results, it is clear the motion generated by the geodesic leads to a motion pattern similar to the demonstrations. Note how the end-effector orientation evolves on the decoded geodesic in the ambient space, showing that the $90$\textdegree~rotation is properly encoded and reproduced.

\subsection{Simulated pouring task}
To evaluate our model on a more complicated scenario, we collected a dataset of pouring task demonstrations on a simulator. The task involves grasping $3$ cans from $3$ different positions and then pouring at $3$ different cups placed at different locations. 
The demonstrated trajectories cross each other, therefore providing a multiple-solution setting. Figure~\ref{fig:Multiple_solution} shows how the geodesic, depicted as a yellow curve, is constructed in the latent space. This geodesic starts from a point on the second demonstration group (blue dots) and switches to the first group (white dots) to get to the given target located in the third group (green dots). As a result, with $3$ sets of demonstrations, all $9$ permutations for grasping any can from the table and then pouring any cup are feasible.

\begin{figure*}
\centering
\begin{subfigure}{.4\textwidth}
  \centering
  \includegraphics[width=\linewidth]{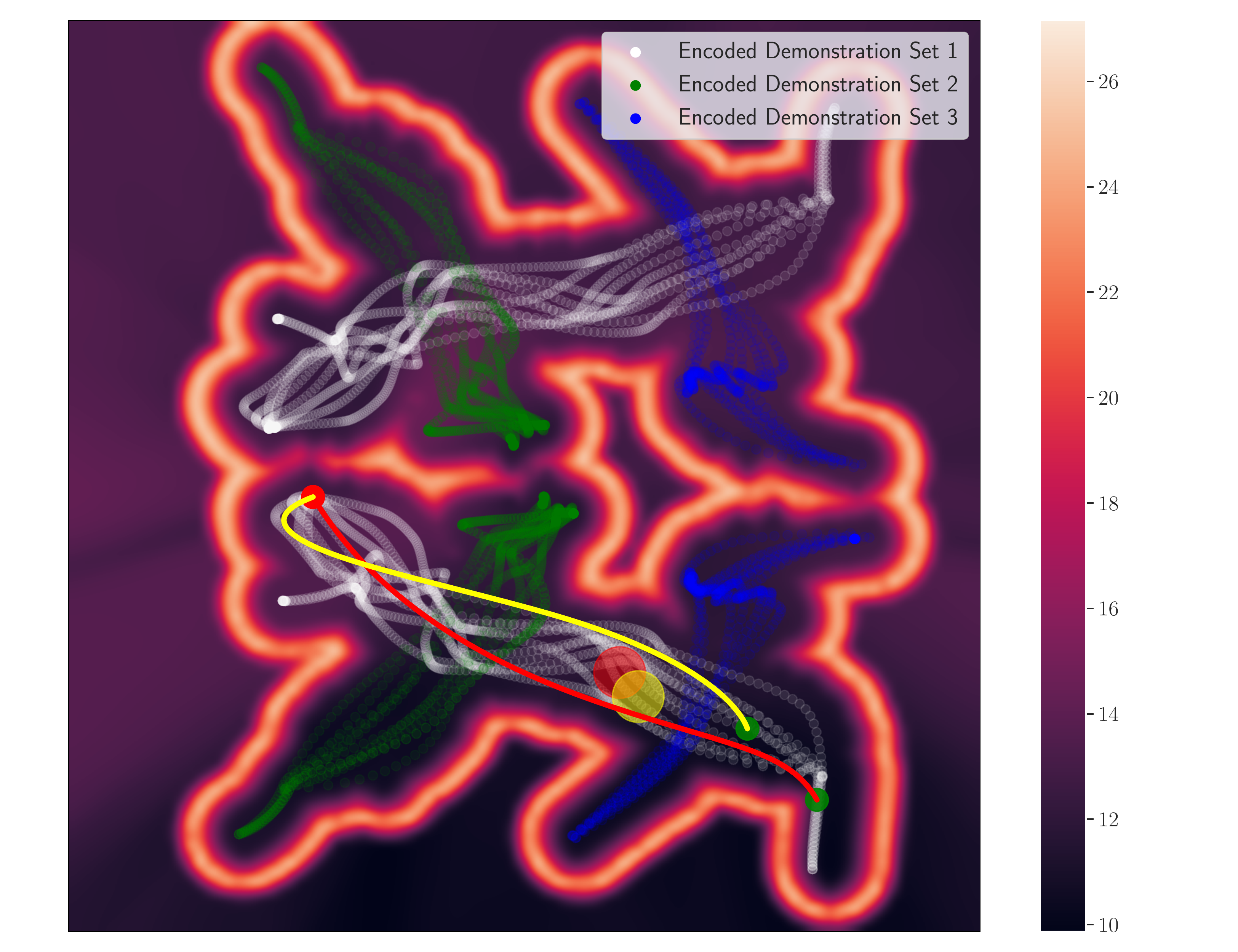}
\end{subfigure}%
\begin{subfigure}{.29\textwidth}
  \centering
  \includegraphics[width=\linewidth]{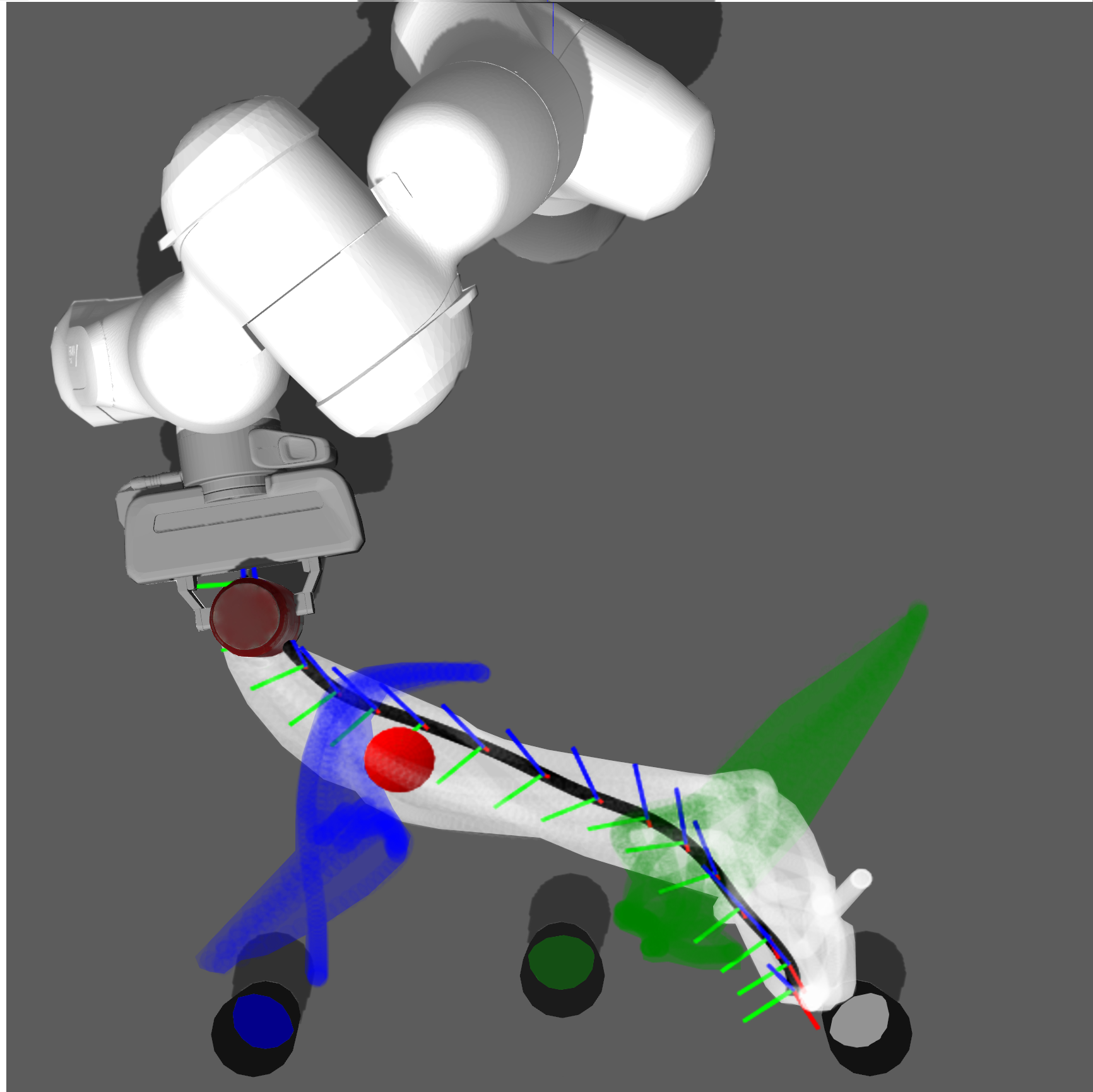}
\end{subfigure}
\begin{subfigure}{.295\textwidth}
  \centering
  \includegraphics[width=\linewidth]{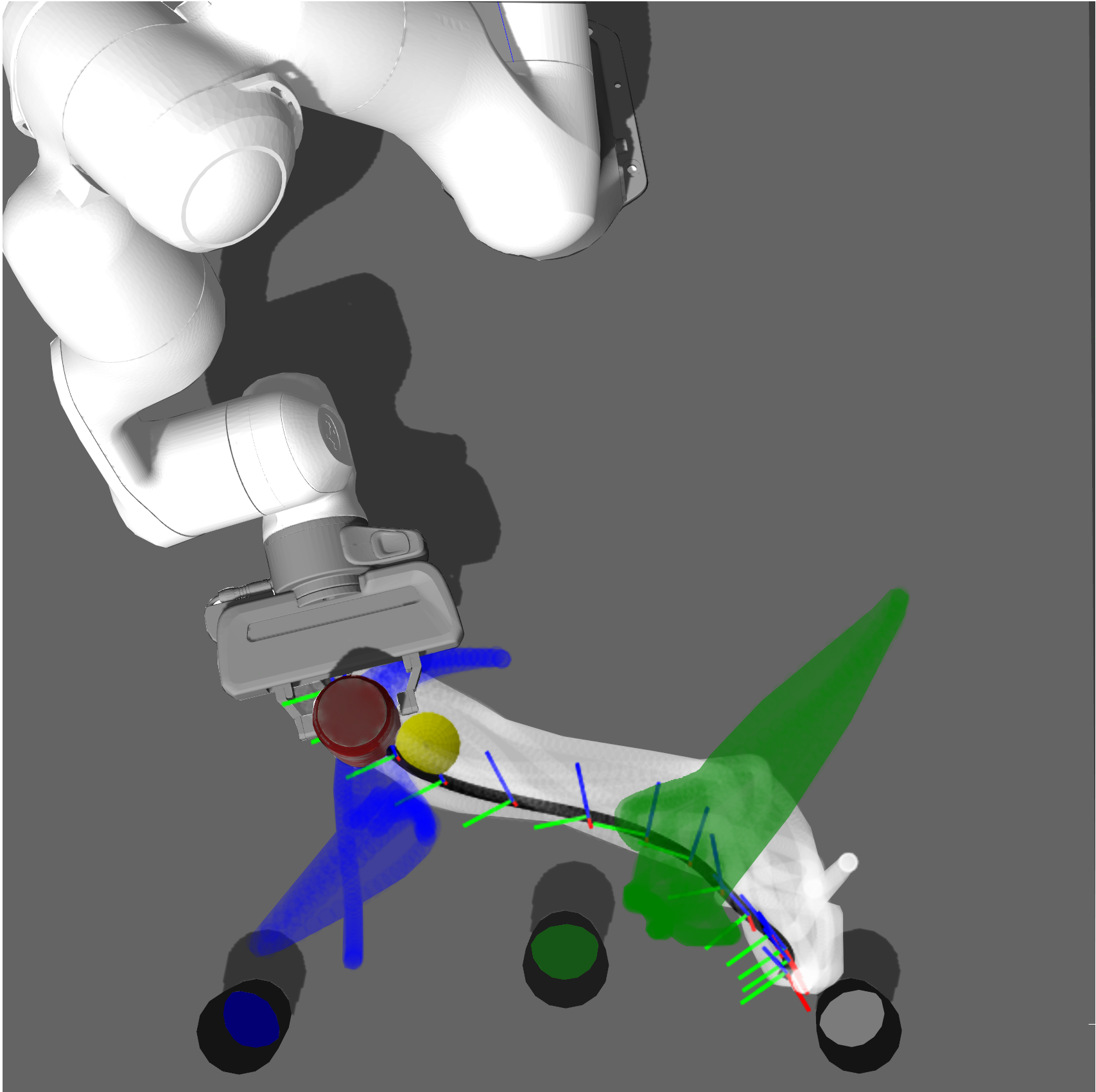}
\end{subfigure}

\caption{\emph{Left}: Red and yellow curves depict geodesics at two different time frames of the same motion, showing how the method computes paths that avoid dynamic obstacles (depicted as red and yellow circles). The background depicts magnification factor derived from the learned manifold. The dot clusters in red, green and blue depict the encoded demonstration sets. \emph{Middle} and \emph{right}: Robot configurations for the two time frames considered in the left plot, from the top perspective in the ambient space. The decoded geodesics are depicted as black curves and the orientations are visualized by coordinate systems.  }
\label{fig:Obstacle_Avoidance}
\end{figure*}

\begin{figure*}
\centering
\begin{subfigure}{.38\textwidth}
  \centering
  \includegraphics[width=\linewidth]{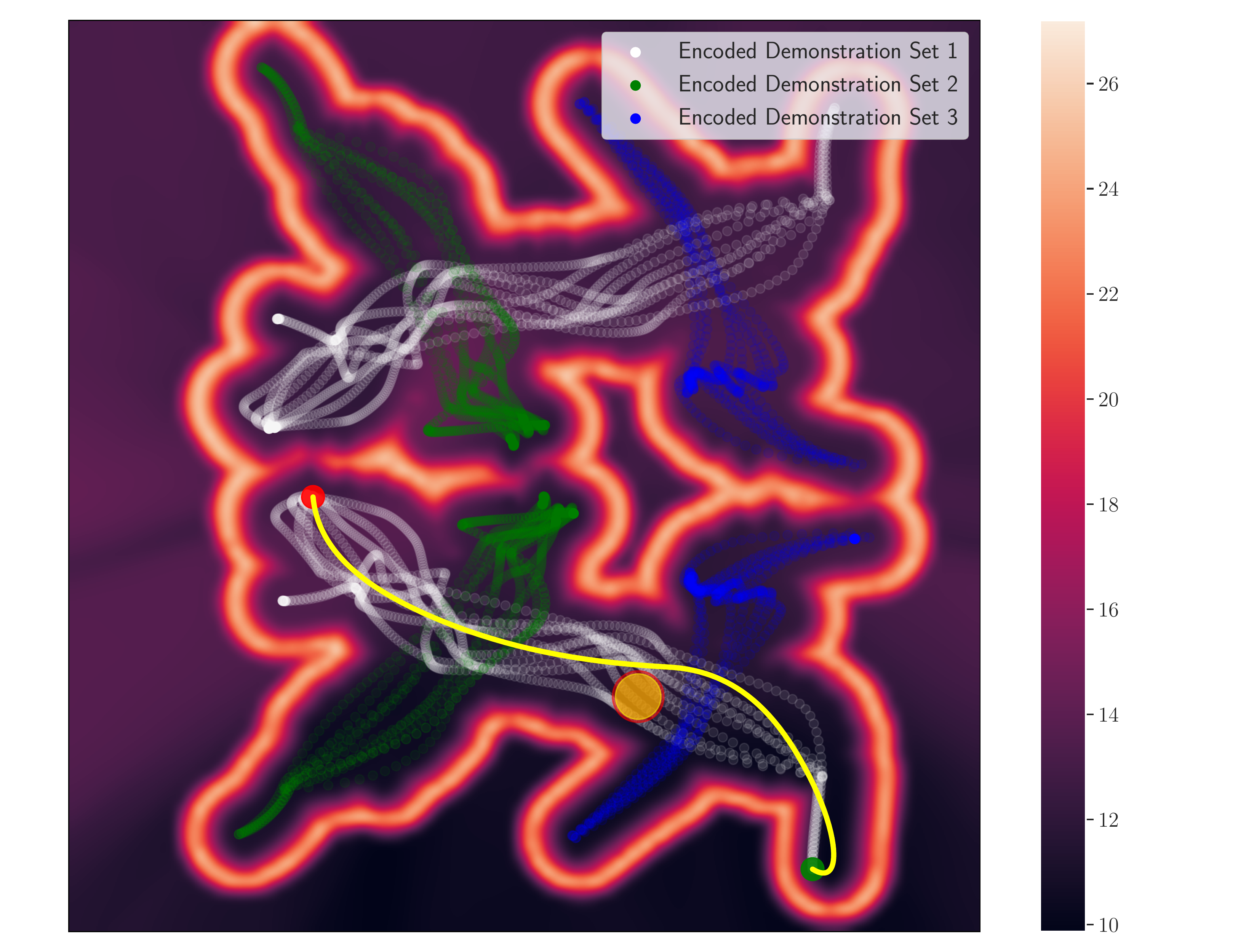}
\end{subfigure}%
\begin{subfigure}{.30\textwidth}
  \centering
 \includegraphics[width=\linewidth]{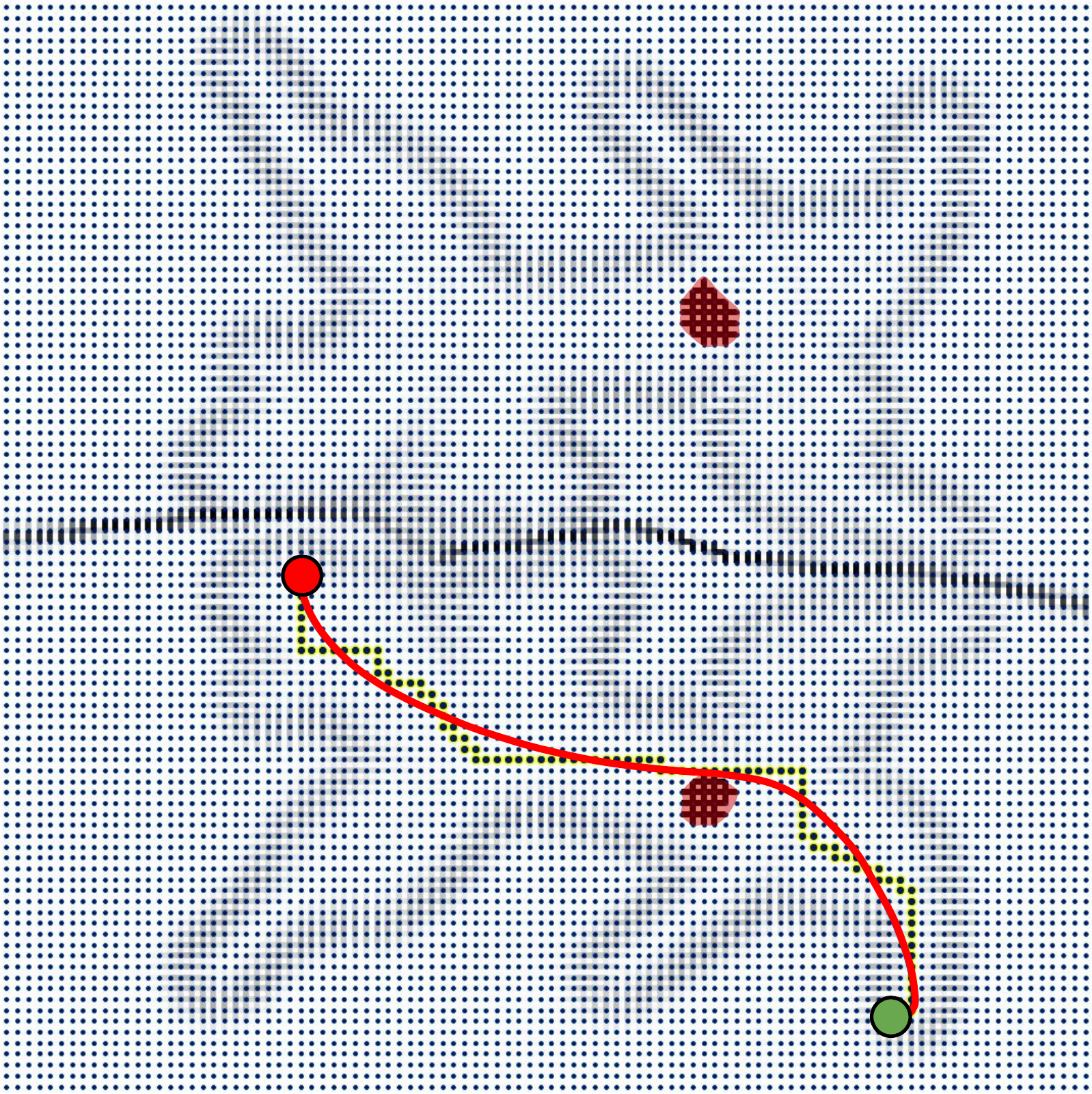}  
\end{subfigure}
\begin{subfigure}{.30\textwidth}
  \centering
  \includegraphics[width=\linewidth]{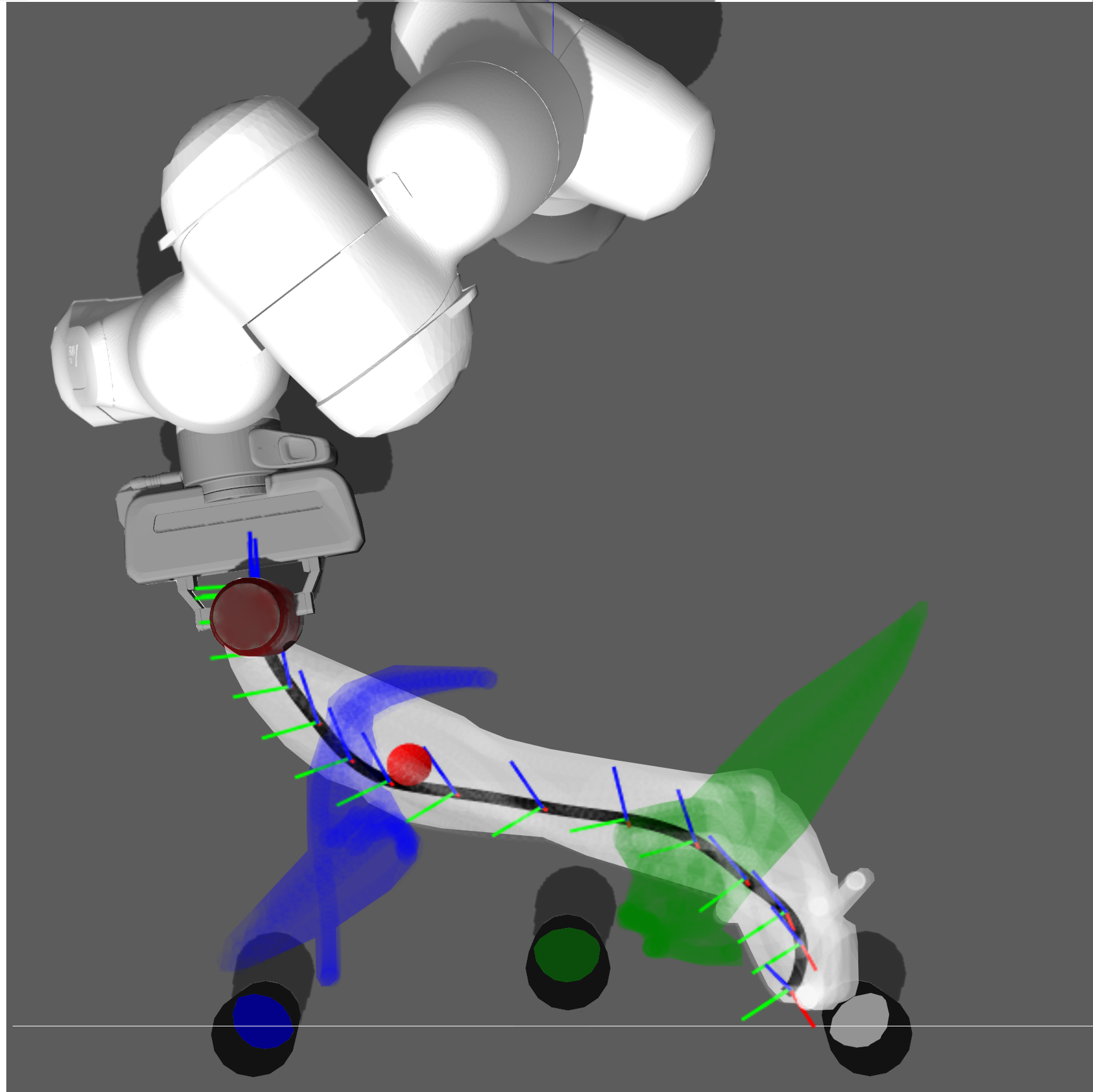}
\end{subfigure}
\caption{\emph{Left}: Geodesic computed on the graph is depicted as yellow curve avoiding the obstacle depicted as yellow circle. The background depicts magnification factor derived from the learned manifold. The dot clusters in red, green and blue depict the encoded demonstration sets. \emph{Middle}: The graph computed from metric learned from data. \emph{Right}: Robot configurations from the top perspective in the ambient space. The decoded geodesic is depicted as a black curve and the orientations are visualized by coordinate systems. }
\label{fig:Discrete_manifold}
\vspace{-5mm}
\end{figure*}
To evaluate the method in the presence of an unseen obstacle, the start and target points of the geodesic path are selected such that it follows one of the demonstration groups. To ensure that the obstacle is in the way, we select its position from the training set. Figure~\ref{fig:Obstacle_Avoidance}-\emph{left} shows the geodesic performing obstacle avoidance while following the geometry of the manifold.
The circular obstacle representation depicted as red and yellow circles in the latent space is just for the sake of visualization. The red and yellow curves represent geodesics avoiding the red and yellow obstacles, correspondingly. These curves correspond to one time frame of the adapted geodesics, showing how our method can deal with dynamic obstacles. The middle and right plots in Fig.~\ref{fig:Obstacle_Avoidance} show the decoded geodesics executed on the simulated robot in the ambient space. The black trajectory shows the decoded geodesic, the dot clusters in red, green, and blue depict the demonstration sets, and the blue sphere depicts the obstacle. The middle plot shows the robot configuration $15$ frames earlier than the plot at the right, displaying the obstacle dynamics during the task execution.  
Figure~\ref{fig:Discrete_manifold}-\textit{middle} illustrates the graph computed from the corresponding data manifold in Fig.~\ref{fig:Discrete_manifold}-\textit{left}. The graph-based geodesic (red curve in Fig.~\ref{fig:Discrete_manifold}-\textit{middle}) is then decoded and executed on the simulated robot (see Fig. ~\ref{fig:Discrete_manifold}-\textit{right}). Although gradient-based and graph-based geodesic computation are both viable options, the latter is faster and thus more suitable for real-time motion generation.

\section{Discussion and conclusions} 
We have proposed a novel LfD approach that learns a Riemannian manifold from human demonstrations and computes geodesics to recover and synthesize learned motion skills. Our proposed geodesic motion generation is capable of planning movements from and to arbitrary points on the data manifold, while avoiding obstacles on the fly. We realize the idea with a variational autoencoder (VAE) over the space of position and orientations of the robot end--effector. Motion is generated with graph--based geodesic computation for real--time motion generation and adaptation. Through extensive evaluation in the simulation, we show geodesic motion generation performs well in different scenarios such as grasping and pouring. 

The proposed methodology can be extended and improved in several directions.
A consequence of learning a manifold skill using VAEs is that data lying outside the manifold may be arbitrarily misrepresented in the latent space $\latent$. Consequently, any reconstruction in the ambient space $\ambient$ may be inaccurate. This may give rise to problems when conditioning on points, e.g.\ new targets, that are located outside the learned manifold. We did not explore this setting in the present paper. One possible solution may involve learning a bijective mapping between old demonstrations and new conditions, and then use this function to transform the learned manifold (e.g., by expanding or rotating) to fit another region of the space.

We introduce dynamical obstacle avoidance as a soft constraint through the ambient metric. If an approach based on hard constraints is to be preferred then one may opt to remove nodes near an obstacle from the graph instead of re-weighting edges. This could provide a computational saving, but one would lose the `complete' Riemannian picture that we find elegant. It would also be straightforward to direct geodesics to go through select \emph{via-points} by slight modifications to the graph algorithm. We did not explore these approaches here.

Our obstacle avoidance formulation only considered simple obstacles at this point, but the strategy can be extended to multiple dynamic obstacles. Instead of working with single Gaussian balls, one can imagine extending the approach to complex obstacle shapes represented as point clouds. This may increase the implementation demands in order to remain real time, but such an extension seems entirely reasonable. It is worth pointing out that to execute a motion safely the obstacle avoidance should be considered for all robot links and not just the end-effector. This requires a more informative manifold that is embedded in the joint space of the robot and an ambient space metric that combines the obstacle information in the Euclidean and joint space simultaneously.



\bibliographystyle{plainnat}
\bibliography{references}
\end{document}